\documentclass[sigconf]{acmart}
\AtBeginDocument{%
  }

\setcopyright{acmlicensed}
\copyrightyear{2018}
\acmYear{2018}
\acmDOI{XXXXXXX.XXXXXXX}
\acmConference[Conference acronym 'XX]{Make sure to enter the correct
  conference title from your rights confirmation email}{June 03--05,
  2018}{Woodstock, NY}
\acmISBN{978-1-4503-XXXX-X/2018/06}

\usepackage{times}  
\usepackage{helvet}  
\usepackage{courier}  
\PassOptionsToPackage{hyphens}{url}
\usepackage{graphicx} 
\usepackage{natbib}  
\usepackage{caption} 
\usepackage{booktabs}
\usepackage{multirow}
\usepackage{makecell}
\usepackage{array}
\usepackage[table]{xcolor}
\usepackage[most]{tcolorbox}
\usepackage{enumitem}
\usepackage{array, xcolor}
\usepackage{tabularx}
\usepackage{listings}

\usepackage{amsmath}
\usepackage{newfloat}
\usepackage{listings}
\DeclareCaptionStyle{ruled}{labelfont=normalfont,labelsep=colon,strut=off} 
\floatstyle{ruled}
\newfloat{listing}{tb}{lst}{}
\floatname{listing}{Listing}
\newcommand{\system}{TripTide}

\title{\system{}: A Benchmark for Adaptive Travel Planning under Disruptions}
\author{Priyanshu Karmakar}
\affiliation{%
  \institution{School of Electrical and Computer Sciences, IIT Bhubaneswar}
  \city{Bhubaneswar}
  \country{India}}
\email{<a25cs09001@iitbbs.ac.in>}

\author{Soumyabrata Chaudhuri}
\authornote{Currently at UT Austin, USA.}
\affiliation{%
  \institution{School of Electrical and Computer Sciences, IIT Bhubaneswar}
  \city{Bhubaneswar}
  \country{India}}
\email{<chaudhurisoumyabrata@gmail.com >}

\author{Shubhojit Mallick}
\affiliation{%
  \institution{Microsoft}
  \country{India}}
\email{<shmallic@microsoft.com>}

\author{Manish Gupta}
\affiliation{%
  \institution{Microsoft}
  \country{India}}
\email{<gmanish@microsoft.com>}

\author{Abhik Jana}
\affiliation{%
  \institution{School of Electrical and Computer Sciences, IIT Bhubaneswar}
  \city{Bhubaneswar}
  \country{India}}
\email{< abhikjana@iitbbs.ac.in>}

\author{Shreya Ghosh}
\affiliation{%
  \institution{School of Electrical and Computer Sciences, IIT Bhubaneswar}
  \city{Bhubaneswar}
  \country{India}}
\email{<shreya.cst@gmail.com >}

\begin{abstract}
Recent efforts like TripCraft and TravelPlanner have advanced the use of Large Language Models (LLMs) for personalized, constraint-aware travel itinerary generation. However, real-world travel rarely unfolds without disruptions. To address this critical gap, we introduce \system{}\footnote{Capturing the ebb and flow of disrupted travel plans}, the first benchmark specifically designed to evaluate LLMs' ability to adapt itineraries in the face of realistic disruptions. 
\system{} models key dimensions such as disruption severity levels and traveler tolerance profiles, enabling nuanced assessment of LLM responses to unexpected events like transit cancellations, weather-related closures, or overbooked attractions. 

We perform a holistic \textcolor{black}{three-fold evaluation. First, we propose}, a suite of novel automatic evaluation metrics, including `Preservation of Intent' (assessing the feasibility and goal retention of revised plans), `Responsiveness' (capturing how promptly and appropriately the LLM addresses disruptions), and a set of `Adaptability' metrics that quantify semantic, spatial, and sequential divergence between original and modified itineraries. \textcolor{black}{Second, we use LLM-as-a-judge approach to evaluate \system{}'s effectiveness in handling disrupted plans. Third, a manual evaluation is conducted to verify whether the revision to address disruptions adequately captures the semantic, spatial, sequential, and responsiveness aspects.} 

Our experiments reveal intriguing mitigation strategies. LLMs tend to maintain \textcolor{black}{strong sequentiality and stable semantic divergence between the original and revised plans, while the spatial deviations are pronounced for shorter plans but diminish for longer ones, suggesting that extended plan length allows better geographic coherence. \system{} also reveals that the LLMs' capability to mitigate the disruptions gradually declines as plan horizons increase.} 
\system{}\footnote{Codebase: \url{https://anonymous.4open.science/r/TripTide-C3A7/}\label{codeFN}} establishes a new foundation for benchmarking real-world robustness in travel planning agents, emphasizing adaptability, personalization, and resilience in LLM-generated itineraries.
\end{abstract}

\usepackage{bibentry}

\begin{document}

\maketitle


\section{Introduction}
Large Language Models (LLMs) have recently been applied to automated travel itinerary generation, with the aim of producing coherent, personalized, and logistically feasible plans utilizing their strong structured reasoning and decision-making abilities. However, despite the increasing use of LLMs for travel itinerary generation, existing works like \textit{TravelPlanner}~\cite{xie2024travelplanner} and \textit{TripCraft}~\cite{chaudhuri2025tripcraft} often assume an ideal, disruption-free environment. Other works, such as \textit{Flex TravelPlanner}~\cite{oh2025flextravelplanner}, have attempted to shed light on the topic, but have failed to capture the real-life disruption scenarios that occur during actual travel planning. As a result, they fail to reflect the true complexity of travel planning in the wild. Therefore, it is essential to study how LLMs respond to travel disruptions (such as flight delays, closure of a point of interest, accommodation constraints, etc.), offering insights into their real-world utility and ability to revise a trip due to disruptions.
    \begin{figure}[t]
        \centering
        \includegraphics[width = 0.50\textwidth]{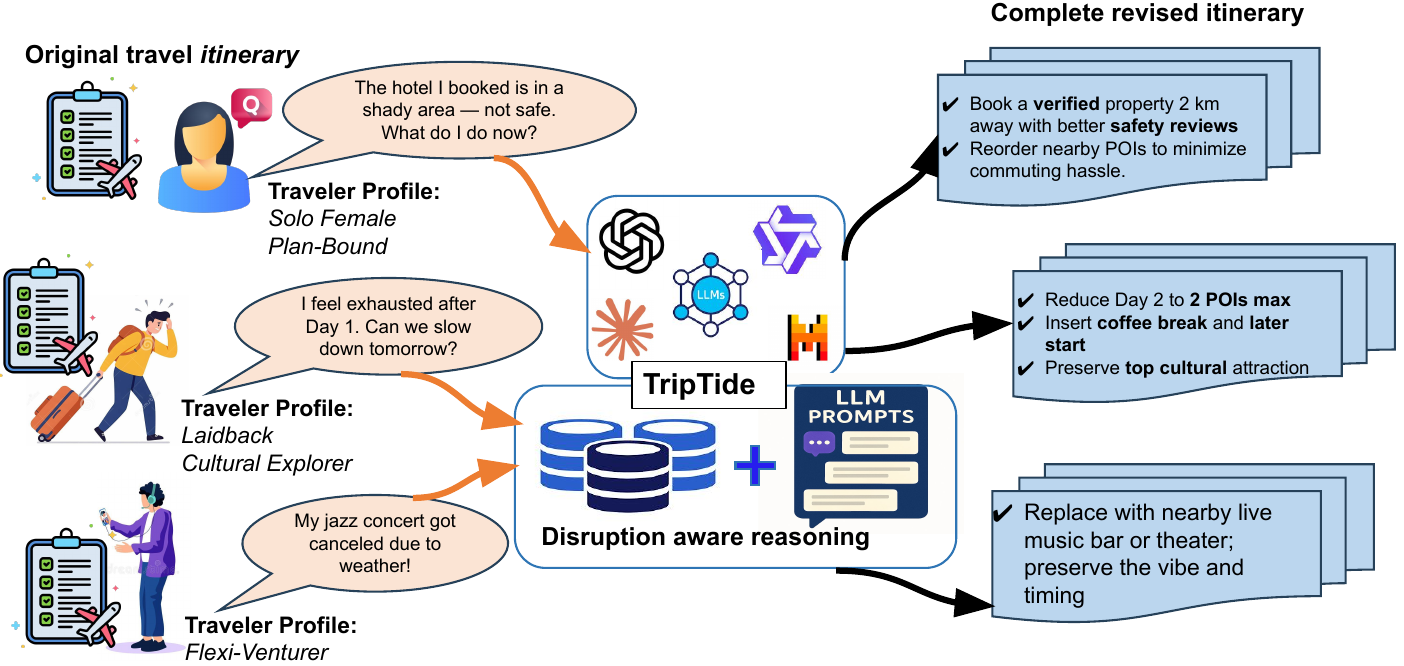}
        \caption{
        Motivating examples of persona-guided itinerary adaptation in TripTide under different types of disruptions
}
        \label{fig:motiv}
    \end{figure}

To fill this critical gap, we augment one of the existing datasets, namely TripCraft \cite{chaudhuri2025tripcraft}, which is composed of 1,000 diverse, constraint-rich travel planning queries and introduce \system{}, a new benchmark that simulates a wide range of realistic travel disruptions (see Fig.~\ref{fig:motiv} for motivating examples of TripTide). These disruptions are crafted with carefully designed metadata, including (a) severity levels: step-level (a single activity is impacted), day-level (an entire day’s plan is affected), and plan-level (multiple days or the overall itinerary requires revision) 
and (b) traveler tolerance profiles (Flexi-Venturer or Plan-Bound). Each disruption instance is grounded in plausible scenarios such as cancelled activities, overbooked accommodations, or weather-related closures, creating a high-fidelity environment to evaluate LLMs' capacity for adaptive planning (See Table~\ref{tab:disruption_types}). This augmented dataset enables us to systematically investigate LLMs' performance in scenarios that require quick thinking and plan revision, with a deep understanding of user preferences.

To assess how effectively LLMs handle travel disruptions, we propose a suite of three novel evaluation metrics tailored to this dynamic context. The `Preservation of Intent' score assesses the extent to which the modified itinerary upholds the essential objectives and experiential priorities of the original plan. The `Responsiveness' score measures the model’s ability to promptly and appropriately respond to the disruption, distinguishing between decisive interventions and vague or evasive modifications. In addition, we introduce a suite of `Adaptability' metrics that quantify the semantic, spatial, and sequential shifts between the original and revised itineraries, thereby offering detailed insights into how the model restructures plans under evolving constraints. \textcolor{black}{We also employ the LLM-as-a-Judge method for automatic evaluation of the generated plans and subsequently validate the experimental results by correlating them with manual evaluations of the disruption-mitigated plans conducted by human experts.}
We found that language models effectively acknowledge disruptions and react to mitigate them; however, they sometimes struggle to maintain narrative coherence between initial and revised plans.

\begin{table}[t]
\centering
\scriptsize
\setlength{\tabcolsep}{3pt}
\begin{tabular}{l p{1.7cm} p{5.2cm}}
\toprule
& \textbf{Subtype} & \textbf{Example / Description} \\
\midrule

\multirow{15}{*}{\rotatebox{90}{\textbf{Disruption Category}}} 
& Transport  & Flight Delays, cancellations, missed connections, baggage issues, strikes, weather, or airport closures \\
& Accommodation  & Poor hygiene, unsafe/remote location, missing promised amenities \\
& Restaurant  & Closures, changes in hours, poor hygiene, no dietary support, overcrowding, long waits, surge pricing \\
& Attractions & Closed for maintenance, weather-related issues, government restrictions, ticket sell-outs, ID rules, overcrowding, limited open days.  \\
& Miscellaneous  & Any event based disruptions like cancellation of events, rescheduling a event, sudden venue changes, event ticket issues, weather or safety risks, also includes any other form of disruptions that has not been categorized like illness, jet lag, fatigue, bad weather, forgotten documents.\\
\midrule

\multirow{3}{*}{\rotatebox{90}{\textbf{Severity}}} 
& Step-Level & Affects an individual activity or segment \\
& Day-Level & Impacts all activities on a particular day \\
& Plan-Level & Necessitates major plan or route overhauls \\
\midrule

\multirow{4}{*}{\rotatebox{90}{\parbox{0.60cm}{\textbf{Traveler Tolerance}}}} 
& Flexi-Venturer & Open to rerouting, substitution, or plan changes \\
& Plan-Bound & Prefers minimal changes and high plan adherence following Disruption severity. \\
\bottomrule
\end{tabular}
\caption{Taxonomy of disruption categories, severity levels, and traveler tolerance profiles}
\label{tab:disruption_types}
\end{table}

Our work makes the following key contributions:
\begin{enumerate}
    \item \textbf{LLM-Based Disruption Response Generation:} We introduce \system{}, the first benchmark specifically designed to evaluate the robustness of LLM in travel planning under realistic disruptions. We develop a \textcolor{black}{few-shot} prompt-driven mechanism to handle different types of disruption by guiding LLMs to generate coherent, minimally altered revised itineraries. 
    \item \textbf{Structured Annotation and Comprehensive Evaluation Metrics:} We contribute structured, annotated, revised plans, curated post-disruption, which serve as a benchmark for evaluating travel planning systems. Additionally, we also propose a set of novel evaluation metrics that measure the preservation of user intent, responsiveness to disruptions and adaptability of the revised plans.
    \item \textbf{Analysis of LLM Planning Behavior under Disruption:} We identify key behavioral patterns in how LLMs handle disruptions, providing insights into their underlying planning heuristics and standard failure modes.
\end{enumerate}

Collectively, these contributions lay the foundation for a new standard in evaluating and improving LLM-driven travel planning systems, particularly in the context of real-world scenarios where disruptions are the norm.

\begin{figure*}[t]
\centering
\includegraphics[width=\linewidth]{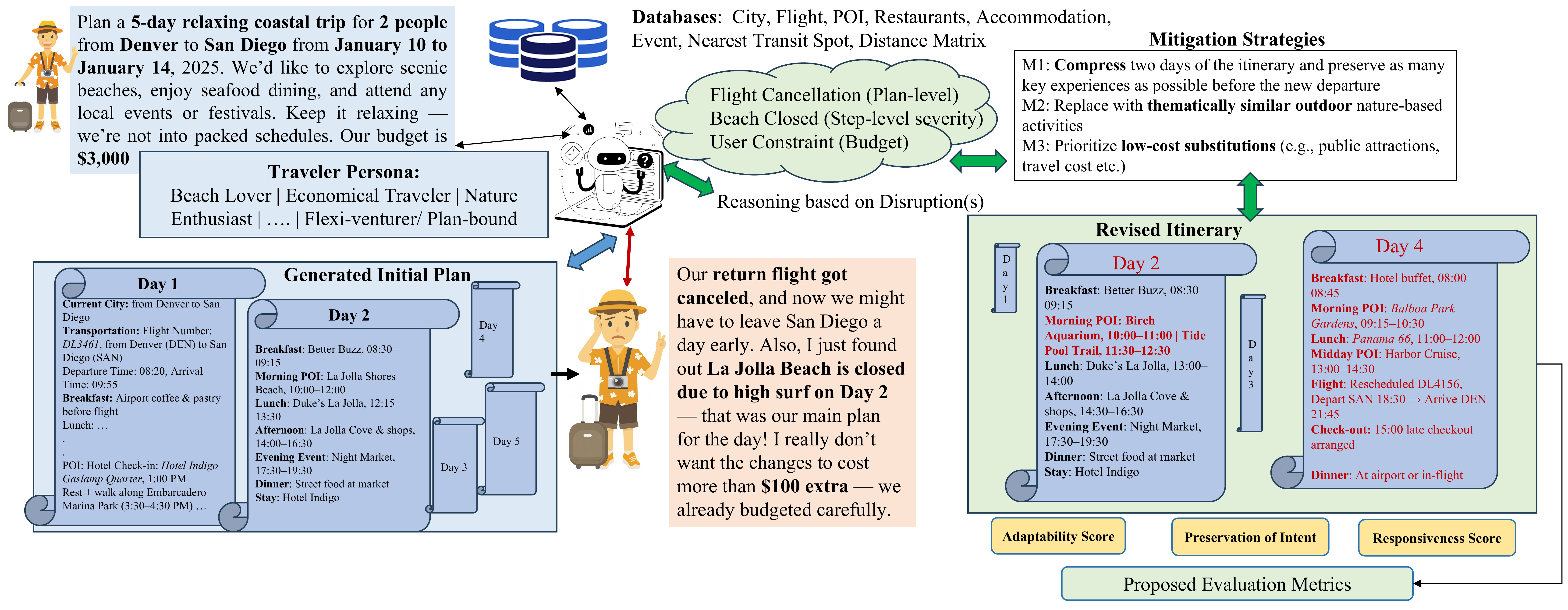}
\caption{Illustration of multi-level disruption handling and adaptive travel replanning by TripTide}
\label{fig:flight-disruption}
\end{figure*}

\section{Related Work}
\textbf{LLMs in Planning tasks.} 
With the advent of LLMs, there has been a growing interest in their ability to perform such planning through natural language~\citep{valmeekam2023planningabilitieslargelanguage, bohnet2024exploringbenchmarkingplanningcapabilities}. LLMs have shown impressive capabilities in task decomposition, commonsense reasoning, and step-by-step decision-making~\citep{wei2022chain, yang2023planning}, making them compelling candidates for real-world planning applications. Recent approaches such as \textit{LLM-Planner}~\citep{song2023llmplanner} enable few-shot grounded planning by iteratively refining high-level plans based on feedback. At the same time, hybrid methods integrating LLMs with classical algorithms like Monte Carlo Tree Search have demonstrated improved planning efficiency~\citep{zhao2024large, swiechowski2023monte}. Other existing works like \textit{TripTailor}~\cite{shen2025triptailorrealworldbenchmarkpersonalized} and~\cite{shao2025chinatravelopenendedbenchmarklanguage} extending the \textit{TravelPlanner} idea talk about the LLM's reasoning capabilities across the China dataset. Despite this progress, LLMs still struggle to reliably generate robust and coherent plans in open-ended environments~\cite{zhang-etal-2025-planning}, often faltering in sub-goal coordination and long-horizon dependencies~\citep{valmeekam2023planningabilitieslargelanguage, kambhampati2023role,singh-etal-2024-personal}.
While chain-of-thought prompting~\cite{wei2022chain} and fine-tuning improve benchmark performance, they falter on novel, complex tasks. Retrieval-based methods such as RAG~\cite{ni2025tpragbenchmarkingretrievalaugmentedlarge} and knowledge graphs~\cite{ijgi13110414, xiao-etal-2020-practice} likewise remain unproven in real-world travel disruptions scenarios.\newline
\noindent\textbf{Travel Planning with LLMs.}
Travel planning poses a multifaceted challenge, involving the coordination of diverse sub-goals such as time management, cost optimization, and personalized user preferences~\citep{gurjar2021should}. The recent application of LLMs to this domain has sparked interest due to their ability to interpret and generate natural language instructions~\citep{xi2025rise, jonnala2025exploring}. Initial benchmarks, such as the one proposed by~\citet{xie2024travelplanner}, highlight the difficulty LLMs face in balancing multiple constraints~\cite{borro2025largelanguagemodelscommonsense}, often producing itineraries that violate feasibility or preference criteria. Although subsequent works~\citep{hao2025largelanguagemodelssolve, gundawar2024robust} report promising results, they typically oversimplify the problem by omitting real-world limitations such as transit schedules, event timing, and dynamic user contexts~\citep{shao2024chinatravel} and using discrete binary evaluation metrics. Although recent efforts~\citep{chaudhuri2025tripcraft, chen2024travelagentaiassistantpersonalized, singh-etal-2024-personal} have advanced the use of continuous metrics to evaluate rationality, personalization, and alignment of user preference in travel plans, they largely overlook the 
disruption-prone nature of real-world travel.

\noindent\textbf{Disruptions in Travel Planning.}
Despite travel being inherently susceptible to disruptions such as weather events, transportation delays, and overbooked attractions, prior studies have largely ignored this aspect of planning. Existing benchmarks and systems assume ideal execution of generated itineraries, without considering how plans should evolve when unexpected obstacles arise. To the best of our knowledge, \system{} is the first benchmark to systematically study how LLMs respond to travel disruptions. 
\system{} simulates a diverse range of real-world disruptions and introduces metrics to assess how well LLMs adapt while preserving user goals and contextual feasibility.

\section{\system{} Dataset Curation}
\system{} consists of 1,000 travel planning queries (adapted from TripCraft) across three trip durations (3-day, 5-day, and 7-day), each paired with a corresponding disruption query and a human-annotated revised plan. These revised plans are designed to respect both the traveler’s persona and disruption tolerance level (e.g., `Flexi-Venturer' or `Plan-Bound'), ensuring that LLMs are evaluated for their ability to maintain personalization and coherence under restrictions. Fig.~\ref{fig:flight-disruption} depicts the block diagram of TripTide illustrating multi-level disruption(s) handling and replanning features of our proposed system. 
\subsection{Disruption Generation}
\subsubsection{Task Overview:} For each of the annotated plans provided in the TripCraft dataset \cite{chaudhuri2025tripcraft}, we generate three potential disruptions per day of the itinerary by prompting GPT-4o \textcolor{black}{and Gemini 2.5 Pro~\cite{comanici2025gemini}.}
Each disruption is then paired with a corresponding disruption query, which is used as input to prompt GPT-4o\footnote{\url{https://openai.com/index/gpt-4o-system-card/}} to generate a persona-aware revised plan. The complete prompts are mentioned in our codebase\footref{codeFN}. Hyperparameter details are provided in Appendix~\ref{app:hyperparams}. 

\subsubsection{Disruption Categories and Scenarios.}
This section outlines the various categories of disruptions that may occur during a trip, along with representative scenarios under each category. These are also illustrated in Table~\ref{tab:disruption_types}.

\noindent{\textbf{1. Flight Disruptions.}} 
These include flight delays, cancellations, missed connections, baggage mishandling, labour strikes, adverse weather conditions, and airport closures.

\noindent{\textbf{2. Accommodation Disruptions.}} 
These can arise due to poor hygiene, unsafe or remote locations, unavailability of promised amenities, or overbooking.

\noindent{\textbf{3. Restaurant Disruptions.}} 
These include unexpected closures, changes in operating hours, poor hygiene standards, inadequate dietary accommodations, overcrowding, lengthy wait times, or surge pricing.

\noindent{\textbf{4. Event and Activity Disruptions.}} 
These include event cancellations, rescheduling, venue changes, ticketing issues, or safety-related concerns due to weather or local advisories.

\noindent{\textbf{5. Miscellaneous Disruptions.}} 
Other disruptions encompass traveler illness, jet lag, fatigue, inclement weather, or forgotten documents, all of which may affect the overall itinerary.

\subsubsection{Disruption Severity Levels.}

As a key feature of our benchmark, we categorize disruptions based on their scope into three levels: \textit{Step-Level}, \textit{Day-Level}, and \textit{Plan-Level}.
    \textbf{Step-Level} disruptions 
    affect a single activity or step within the itinerary (e.g., a visit to a specific attraction or a meal reservation).
    \textbf{Day-Level} disruptions 
    impact the set of activities scheduled for a particular day, requiring partial rescheduling or substitution of that day's Point-of-Interest (POI) list.
    \textbf{Plan-Level} disruptions 
    have broader consequences, potentially requiring significant changes across multiple days of the itinerary, including complete reordering or replacement of the POI list.

\subsubsection{User Personas and Tolerance Levels.} 

We categorize travelers using structured user personas that encapsulate their goals, preferences, and behavioral traits during a trip. \textcolor{black}{Examples of such personas include the \textit{Adventure Seeker}, the \textit{Cultural Explorer}, the \textit{Economical Traveler}, or the \textit{Mountain Enthusiast}, each reflecting distinct travel styles and motivations.}
Each persona in our framework is further associated with a predefined tolerance level to disruption, ranging from highly flexible to strictly structured, based on factors such as adaptability, planning preferences, and sensitivity to change. We introduce a novel, first-of-its-kind modeling of disruption tolerance in travel planning, distinguishing between two representative user types: the \textit{Flexi-Venturer}, who readily adapts to changes, and the \textit{Plan-Bound} traveler, who prefers minimal deviation from the original itinerary. 
\textcolor{black}{While the user persona reflects the traveler’s preferences, such as whether they seek relaxation or adventures, the disruption tolerance level indicates how well the traveler adapts to disruptions. Incorporating the tolerance level alongside the user persona enables a more accurate representation of travelers’ real-world behaviors under disruptions.}

\noindent{\textbf{Flexi-Venturer Traveler.}}  
This traveler exhibits high adaptability and is open to changes in the itinerary following a disruption, regardless of its scope. They are comfortable with dynamic re-planning and show low sensitivity to deviations from the original plan.\\
\noindent{\textbf{Plan-Bound Traveler.}}  
In contrast, the Plan-Bound traveler is rigid and prefers to adhere strictly to the original itinerary. Their acceptance of changes is constrained by the disruption severity. For instance, in the case of a step-level disruption, they permit only that specific step to be modified. 


\subsection{Annotation Construction and Quality Control}

\subsubsection{Human-Level Annotation and Disruption Query Generation.}

To ensure realistic modeling of disruptions in travel itineraries, we adopt a rigorous human-in-the-loop annotation framework. For each day of travel, we identify three plausible disruptions, resulting in a comprehensive pool of disruption scenarios per plan. Specifically, for the 344 three-day plans, this yields 2489 disruption cases. Similarly, disruptions are generated for all five-day and seven-day plans using the same per-day logic. Table~\ref{tab:disruption_summary} contains the detailed distribution per disruption category. In total, our dataset includes \textbf{11,058} possible disruptions spanning a diverse range of conditions, traveler profiles, and destinations.

Human annotators\footnote{Undergraduate interns at our lab} were then tasked with sampling the most contextually meaningful disruption for each plan. They carefully reviewed the full reference itinerary and selected disruptions that emphasized realism, diversity (in terms of type and severity), and traveler-specific relevance. For each selected disruption, annotators collaborated with the GPT-4o and Gemini 2.5 Pro models to generate natural language queries that emulate how travelers would express real-world concerns (e.g., \textit{``My flight's been delayed. What now?''}). 
Annotators then revised the original itinerary in response to the disruption, ensuring logical adaptation, temporal feasibility, and overall trip coherence. 
That helped to encode grounded, expressive, and context-rich examples that reflect how disruptions are perceived and handled in real-world travel. Detailed annotation guidelines are in Appendix~\ref{app:humanAnnotation}.

\begin{table}[htbp]
\centering
\scriptsize
\setlength{\tabcolsep}{5pt}
\begin{tabular}{llccc r}
\toprule
\textbf{Constraint Dimension} & \textbf{Subtype} & \multicolumn{3}{c}{\textbf{Plan Duration}} & \textbf{Total} \\
\cmidrule(lr){3-5}
& & \textbf{3-day} & \textbf{5-day} & \textbf{7-day} & \\
\midrule

\multirow{6}{*}{\textbf{Disruption Category}} 
& Transport     & 118  & 364  & 555  & 1,037 \\
& Accomodation  & 225  & 416  & 418  & 1,059 \\
& Restaurant    & 1,072 & 1,223 & 1,618 & 3,913 \\
& Attractions   & 980  & 1,223 & 1,618 & 3,821 \\
& Miscellaneous & 94   & 437  & 697  & 1,228 \\
& \textbf{Total} & \textbf{2,489} & \textbf{3,663} & \textbf{4,906} & \textbf{11,058} \\
\midrule

\multirow{4}{*}{\textbf{Disruption Severity}} 
& Step-level    & 111  & 140  & 134  & 385 \\
& Day-level     & 117  & 111  & 131  & 359 \\
& Plan-level    & 116  & 73   & 67   & 256 \\
& \textbf{Total} & \textbf{344} & \textbf{324} & \textbf{332} & \textbf{1,000} \\
\midrule

\multirow{3}{*}{\textbf{Traveler Tolerance}} 
& Flexi-Venturer & 172 & 162 & 166 & 500 \\
& Plan-Bound      & 172 & 162 & 166 & 500 \\
& \textbf{Total} & \textbf{344} & \textbf{324} & \textbf{332} & \textbf{1,000} \\
\bottomrule
\end{tabular}
\caption{Distribution of disruptions by category, severity, and traveler tolerance across different plan durations.}
\label{tab:disruption_summary}
\end{table}

\subsubsection{Automated Script-Based Verification.}

To complement human annotations with structural and semantic correctness, we implemented a robust script-based validation pipeline. The primary goal was to ensure that all entities mentioned in the revised plan, such as accommodations, flights, events, and attractions, were strictly derived from the original reference itinerary. The script conducts thorough entity alignment checks and flags any hallucinated or mismatched elements that are not grounded in the source data.

Additionally, the script ensures logical consistency across time, location, and activity sequences. It verifies that there are no temporal conflicts (e.g., double-booked time slots), that inter-city transitions are feasible, and that each revised plan respects the budget and structural constraints of the original. This automated check serves as a second layer of quality assurance, substantially reducing annotation noise. Together, our three-tiered curation strategy, combining expert human annotation with rule-based verification, \textcolor{black}{followed by LLM-as-a-Judge evaluation alongside assessments from human experts}, ensures a high-integrity dataset for benchmarking disruption-aware travel planning systems.

\subsubsection{Disruption Categorization and Tolerance Profiling.}
To model user preferences and plan flexibility, each disruption is categorized into a level, whereas the traveler is tagged with a tolerance level, reflecting how tolerant a traveler might be toward changes. Table~\ref{tab:disruption_summary} shows the distribution of tolerance levels and disruption severity across different plan durations.

\section{Automated Evaluation}
\label{sec:automated}
\subsection{Evaluation Metrics}
We propose evaluation along three axes: \textit{preservation of intent}, \textit{responsiveness}, and \textit{adaptability}. Given inputs (current plan, disruption, disruption severity, and disruption tolerance), the model generates an alternative feasible plan. The following novel metrics quantify the effectiveness of this new plan.\\

\noindent\textbf{1. Preservation of Intent.}  
This metric evaluates whether the user’s original travel intent, as characterized by constraints and preferences, is retained in the revised plan. We apply the commonsense and hard constraint pass rates (CPR and HCPR), delivery rate, and final pass rate introduced by TravelPlanner \cite{xie2024travelplanner} to the final alternative plan to estimate the percentage of intent preserved. Higher values are preferred.

\noindent\textbf{2. Responsiveness.}  
This measures whether the model acknowledges the disruption and makes meaningful alterations to the original plan.
For each day, the responsiveness rate is given by:
\begin{equation}
\small
\text{Responsiveness Rate} = \frac{\text{\#Mitigated Plans}}{\text{\#Total Plans}}
\end{equation}
Higher values are preferred.

\noindent\textbf{3. Adaptation Quality.}  
This measures the quality of adaptation, specifically, whether the revised plan offers a realistic and contextually appropriate substitute or reorganization. It evaluates if the model suggests thematically consistent alternatives while minimizing disruptions. Each score is finalized by taking the absolute difference between the initial plan and the revised plan score. Thus, a lower value of the metric score reflects that the revised plan is more thematically consistent with minimal changes.  The score is decomposed into three components: semantic, spatial, and sequential, each measured as follows: 

\begin{table*}[t]
\centering
\scriptsize
\setlength{\tabcolsep}{2.6pt}
\renewcommand{\arraystretch}{1.15}
\begin{tabular}{
p{0.100\textwidth} 
p{0.055\textwidth} 
>{\centering\arraybackslash}p{0.07\textwidth} 
>{\centering\arraybackslash}p{0.055\textwidth} 
>{\centering\arraybackslash}p{0.055\textwidth} 
>{\centering\arraybackslash}p{0.055\textwidth} 
>{\centering\arraybackslash}p{0.055\textwidth} 
>{\centering\arraybackslash}p{0.07\textwidth}  
>{\centering\arraybackslash}p{0.085\textwidth} 
>{\centering\arraybackslash}p{0.085\textwidth} 
>{\centering\arraybackslash}p{0.085\textwidth} 
>{\centering\arraybackslash}p{0.085\textwidth} 
}
\toprule
\multirow{2}{*}{\textbf{Model Name}} &
\multirow{2}{*}{\makecell{\textbf{Plan}\\ \textbf{Duration}}} &
\multirow{2}{*}{\makecell{\textbf{Delivery}\\ \textbf{Rate (\%)}$\uparrow$}} &
\multicolumn{2}{c}{\textbf{CPR (\%)}$\uparrow$} &
\multicolumn{2}{c}{\textbf{HCPR (\%)}$\uparrow$} &
\multirow{2}{*}{\makecell{\textbf{Final Pass}\\ \textbf{Rate (\%)}$\uparrow$}} &
\multicolumn{3}{c}{\textbf{Adaptability (\%)}} &
\multirow{2}{*}{\makecell{\textbf{Responsive-}\\ \textbf{ness Rate (\%)}$\uparrow$}} \\
\cmidrule(lr){4-5}\cmidrule(lr){6-7}\cmidrule(lr){9-11}
& & & \textbf{Micro} & \textbf{Macro} & \textbf{Micro} & \textbf{Macro} &
& \makecell{\textbf{Semantic Score}\\ ($A_{\text{sem}}$)$\downarrow$} &
  \makecell{\textbf{Spatial Score}\\ ($A_{\text{spa}}$)$\downarrow$} &
  \makecell{\textbf{Sequential Score}\\ ($A_{\text{seq}}$)$\downarrow$} & \\
\midrule
\multirow{3}{*}{GPT-4o}
& 3-day & 100.00 & 91.02 & 29.95 & 50.93 & 48.84 & 29.37 & 0.34 & 1.22 & 10.89 & 89.53 \\
& 5-day & 99.69 & \textbf{91.46} & \textbf{43.21} & \textbf{67.38} & \textbf{67.59} & \textbf{41.98} & \textbf{0.02} & 0.66 & \textbf{1.86} & 81.79 \\
& 7-day & 100.00 & 87.35 & 33.14 & 55.59 & 57.54 & 32.54 & 0.20 & \textbf{0.30} & 5.58 & 79.82 \\
\midrule
\multirow{3}{*}{Qwen2.5-7B-Instruct}
& 3-day & 100.00 & 89.56 & 31.16 & 54.02 & 51.93 & 29.98 & 0.67 & 5.94 & 56.26 & 97.67 \\
& 5-day & 100.00 & 89.05 & 29.42 & 58.77 & 59.76 & 29.12 & 0.04 & 1.25 & 28.46 & 82.02 \\
& 7-day & 97.54 & 80.98 & 16.26 & 25.16 & 29.15 & 16.27 & 1.24 & 2.10 & 24.04 & 88.67 \\
\midrule
\multirow{3}{*}{Phi-4 mini Instruct}
& 3-day & 45.23 & 39.75 & 18.68 & 20.36 & 24.49 & 18.26 & 26.27 & 3.32 & 26.64 & 75.94 \\
& 5-day & 83.14 & 68.81 & 6.18 & 18.89 & 13.28 & 5.87 & 7.01 & 1.45 & 30.13 & 50.31 \\
& 7-day & 100.00 & 73.43 & 3.06 & 7.67 & 7.96 & 3.06 & 30.34 & 27.78 & 42.74 & \textbf{100.00} \\
\bottomrule
\end{tabular}
\caption{Metric scores for preservation of user intent (Delivery, CPR, HCPR, Final Pass), Adaptability ($A_{\text{sem}}$, $A_{\text{spa}}$, $A_{\text{seq}}$), and Responsiveness across models and plan durations. Differences are absolute and scaled to percentage points.}
\label{tab:intent-adapt}
\end{table*}

\subparagraph{i) Semantic Adaptability ($A_{\text{sem}}$).}  
This component measures how thematically consistent the revised plan is with the original plan. Specifically, it quantifies the semantic closeness between PoIs in the initial and revised itineraries with the user persona using BERT-based cosine similarity. While we adopt the persona score formulation introduced in \citep{chaudhuri2025tripcraft}, in our setting, it is repurposed to reflect plan-level thematic continuity rather than alignment with the user persona:
\begin{equation}
\small
\bar S_{\text{ps}} = \frac{1}{M \cdot N} \sum_{j=1}^{M} \sum_{i=1}^{N} \frac{\mathbf{p}_j \cdot \mathbf{q}_i}{\|\mathbf{p}_j\| \|\mathbf{q}_i\|}
\end{equation}
where \( \mathbf{p}_j \) and \( \mathbf{q}_i \) are BERT embeddings of the $j$-th persona component and $i$-th PoI name respectively, \(M\) is the number of persona components, and \(N\) is the number of PoIs in the plan. The semantic adaptability is then:
\begin{equation}
\small
A_{\text{sem}} = \left| \bar S_{\text{ps}}^{\text{i}} - \bar S_{\text{ps}}^{\text{r}} \right|
\end{equation}
where $\bar S_{\text{ps}}^{\text{i}}$ and $\bar S_{\text{ps}}^{\text{r}}$ are the persona scores of the initial and revised plans, respectively. Lower values are preferred.

\subparagraph{ii) Spatial Adaptability ($A_{\text{spa}}$).}  
This measures the difference in spatial convenience between the original and revised plans. Following \cite{chaudhuri2025tripcraft}, we compute the spatial score for each plan using:
\begin{eqnarray}
\small
S_s(d) =
\begin{cases}
1 - 0.5 \left( \frac{d}{d_0} \right), & \text{if } d \leq d_0 \\
0.5 \exp \left( -\lambda (d - d_0) \right), & \text{if } d > d_0
\end{cases}\\
\bar{S}_{\text{spatial}} = \frac{1}{N} \sum_{i=1}^{N} S_s(d_i)
\end{eqnarray}
where \( d_i \) is the distance of the $i$-th PoI from the nearest public transit station, \(d_0\) is the distance threshold, and \( \lambda\) is the decay rate. Therefore, spatial adaptability is:
\begin{equation}
\small
A_{\text{spa}} = \left| \bar S_{\text{spatial}}^{\text{i}} - \bar S_{\text{spatial}}^{\text{r}} \right|
\end{equation}
where $\bar S_{\text{spatial}}^{\text{i}}$ and $\bar S_{\text{spatial}}^{\text{r}}$ are the spatial scores of the initial and revised plans, respectively. Lower values are preferred.

\subparagraph{iii) Sequential Adaptability ($A_{\text{seq}}$).}  
This evaluates changes in the order of PoIs across days using the normalized edit distance:
\begin{equation}
\small
A_{\text{seq}} = \frac{\text{ED}(\mathcal{G}, \mathcal{A})}{\max(|\mathcal{G}|, |\mathcal{A}|)}
\end{equation}
where \( \mathcal{G} \) and \( \mathcal{A} \) are PoI sequences of the initial and revised plans, and \( \text{ED}(\cdot, \cdot) \) is the edit distance. We average this over all days of both the plans to get \( A_{\text{seq}}\). Lower values are preferred.

\subsection{Results}

We report the performance\footnote{We evaluate the models on the entire dataset due to the absence of a predefined train-test split. These evaluations are diagnostic in nature and are intended to inform design insights rather than generalization performance.} of GPT-4o and Qwen2.5-7B-Instruct and Phi-4-mini Instruct across initial and revised itineraries using \textit{Preservation of Intent} metrics, \textit{Adaptability} metrics (\textit{semantic}, \textit{spatial}, \textit{sequential}), and \textit{Responsiveness} in Table~\ref{tab:intent-adapt} across three plan durations: 3-day, 5-day, and 7-day trips. \\
\noindent\textbf{Preservation of Intent. } 
Both GPT-4o and Qwen2.5-7B-Instruct achieve high delivery rates ($\geq$99.69\%) for 3-day and 5-day plans, with a slight drop observed for Qwen2.5-7B-Instruct at 7-day (97.54\%), \textcolor{black}{whereas Phi-4 mini struggles to deliver for shorter travel plans but provides a perfect score for 7-day (100.00\%)}. Across commonsense (CPR) and hard constraint (HCPR) metrics, GPT-4o consistently achieves moderate CPR and HCPR micro scores, with the best performance on 5-day plans. Qwen-7B-Instruct shows comparable or slightly higher commonsense compliance for shorter durations, but its hard constraint adherence deteriorates sharply in 7-day plans (HCPR micro drops to 25.16\%). \textcolor{black}{On the other hand, for Phi-4 mini Instruct, CPR is uneven, i.e., micro improves with duration (39.75→68.81→73.43) while macro collapses (18.68→6.18→3.06); HCPR stays low and declines (micro 20.36→18.89→7.67, macro 24.49→13.28→7.96).} This suggests that Qwen2.5-7B-Instruct and Phi4-mini-Instruct's constraint reasoning ability diminishes with increasing plan complexity, while GPT-4o maintains more stable performance. 
\textcolor{black}{In terms of final pass rates, GPT-4o leads overall: peaking on 5-day plans at 41.98\% (29.37\% for 3-day, 32.54\% for 7-day). Qwen2.5-7B-Instruct remains around $\sim$30\% on 3- and 5-day plans (29.98\% and 29.12\%) but drops markedly to 16.27\% on 7-day, while Phi-4 mini Instruct is consistently low (18.26\%→5.87\%→3.06\%).}
These trends emphasize GPT-4o’s, Qwen2.5-7B-Instruct's and Phi-4-mini Instruct's reliability in producing plans, while also underscoring the need for improved mechanisms to ensure comprehensive constraint satisfaction.\\
\noindent\textbf{Semantic Adaptability ($A_{\text{sem}}$). } GPT-4o maintains strong semantic fidelity across all plan durations. The differences between initial and revised semantic scores are low (lower is better): 0.34 (3-day), 0.02 (5-day), and 0.20 (7-day), indicating minimal semantic drift post-revision. The near-zero shift for 5-day plans suggests a precise understanding of user intent, even under revision constraints. 
Qwen2.5-7B-Instruct also achieves consistently low semantic drift across all durations: 0.67 (3-day), 0.04 (5-day), and 1.24 (7-day). These results indicate strong alignment with user personas during plan revisions, on par with GPT-4o. The slightly higher shift for 7-day plans could be attributed to inherent possibility of higher drifts in re-planning when facing disruptions in longer itineraries.
\textcolor{black}{While Phi4-mini-Instruct results show decreasing semantic coherence between original and revised plan, where the semantic scores are as high as 30.34\%. This indicates that the Phi-4-mini struggles to understand the user's intent under disruptions. }\\
\noindent\textbf{Spatial Adaptability ($A_{\text{spa}}$).} For GPT-4o, the spatial score remains low (lower is better) across revisions: 1.22 (3-day), 0.66 (5-day), and 0.30 (7-day), indicating minimal spatial drift between the original and revised plans, with the smallest drift for longer (7-day) itineraries. These indicate that GPT-4o consistently restructures plans to group spatially adjacent entities after disruption. The smaller drift for longer duration plans supports the hypothesis that the model leverages spatial flexibility more readily in high-duration settings.
On the other hand, Qwen2.5-7B-Instruct leads to higher spatial drift across all durations: 5.94 (3-day), 1.25 (5-day), and 2.10 (7-day). These results indicate that Qwen2.5-7B-Instruct leads to significant spatial score changes compared to GPT-4o. Phi-4-mini-Instruct shows modest drift for shorter plans (3-day: 3.32; 5-day: 1.45) but fails to maintain spatial coherence for 7-day, with a score of 27.78, indicating difficulty scaling to longer itineraries. \\
\noindent\textbf{Sequential Adaptability (\(A_{\text{seq}}\)).}  
Sequential adaptability measures the average change in PoI ordering across days, based on normalized edit distance between the initial and revised plans. A lower \(A_{\text{seq}}\) implies lower disruption to the temporal structure. 
As presented in Table~\ref{tab:intent-adapt}, GPT-4o exhibits consistently low sequential scores (lower is better): 10.89 (3-day), 1.86 (5-day), and 5.58 (7-day). These scores indicate minimal reordering of PoIs post-disruption, particularly in 5-day itineraries, suggesting that GPT-4o prioritizes original sequence structure while performing disruption constraint satisfaction. In contrast, Qwen2.5-7B-Instruct shows significantly higher \(A_{\text{seq}}\) scores: 56.26 (3-day), 28.46 (5-day), and 24.04 (7-day), indicating much weaker alignment with the original temporal order. While the score decreases with plan length, due to accumulating complexity, Qwen2.5-7B-Instruct remains relatively less sequence-preserving than GPT-4o. \textcolor{black}{Similarly, Phi-4-mini instruct also shows higher sequential scores: 26.64 (3-day), 30.13 (5-day) and 42.74 (7-day). The longer the plan horizon, the higher the sequential score. This indicates the model performs poorly in maintaining the sequentiality of the POI list when the plan horizon increases.}
This contrast highlights a key trade-off: Qwen2.5-7B-Instruct's and Phi-4-mini Instruct's aggressive re-optimization often disrupts sequence, while GPT-4o adopts a more conservative, order-aware strategy. \\
\noindent\textbf{Responsiveness Rate.} Responsiveness, defined as the proportion of disruption-handled plans, drops from 89.53\% (3-day) to 81.79\% (5-day) and further to 79.82\% (7-day). This monotonic decline suggests that GPT-4o faces increasing difficulty as plan complexity and temporal scope increase. Despite strong semantic and spatial preservation, disruption resolution becomes harder to maintain across extended itineraries.
Qwen2.5-7B-Instruct follows the same downward trend with increasing duration from 3-day to 5-day but an increase for 7-day: 97.67\% (3-day), 82.02\% (5-day), and 88.67\% (7-day). Interestingly, Qwen2.5-7B-Instruct achieves a higher responsiveness rate in longer plans compared to GPT-4o, despite more aggressive spatial edits, which indicates better disruption handling at scale.
\textcolor{black}{The phi-4-mini instruct model gives again surprising results while checking for mitigation nature of the revised plan. The models gives 100\% responsiveness rate for 7-day travel planning, while the rate for 3-day and 5-day are low.} \\
\noindent\textbf{Summary.} GPT-4o demonstrates robust semantic preservation and modest spatial reorganization capacity. Its ability to mitigate disruptions decreases slightly with longer durations, and its static sequential performance reflects a need for stronger modeling of temporal dependencies.
 Qwen2.5-7B-Instruct demonstrates stronger responsiveness in longer itineraries than GPT-4o.
 \textcolor{black}{Phi-4-mini often fails to deliver accurate plans when disruptions occur, but it correctly identifies the disruption. Additional result analysis is in Appendix ~\ref{app:resultsAnalysis}.}
 
\section{LLM-as-a-Judge based Evaluation}
\textcolor{black}{In this section, we present an orthogonal, Llama-as-judge perspective on GPT-4o revised plans, summarizing perceived quality across 3/5/7-day itineraries.}

\subsection{Experimental Setup}
\textcolor{black}{In addition to the novel evaluation metrics discussed in Section~\ref{sec:automated}, we further analyze the generated travel itineraries using a separate LLM. Motivated by the idea of \textit{LLM-as-a-Judge}~\cite{zheng2023judging}, we adapt our prompt and evaluate the responses accordingly. As shown in Table ~\ref{tab:intent-adapt}, GPT-4o consistently achieves the best average performance across our evaluation metrics. Hence, we perform this evaluation against GPT-4o outputs.} 

\textcolor{black}{
For each user query, we provide the original plan, the disruption information, and the revised plan generated by the model. We then rate the quality of the revised plan on a scale of 1 to 5, where a score of 1 indicates the worst case (such as when the regenerated plan is nearly identical to the original or fails to meet the user persona's needs) while 5 indicates the best case (disruption is fixed fully). The detailed scoring scheme is presented in Table~\ref{tab:judge_rubric}. 
We perform this evaluation using the Llama-3.1-8B-Instruct\footnote{https://huggingface.co/meta-llama/Llama-3.1-8B} LLM. By evaluating GPT-4o’s responses in this manner, and by applying a carefully designed evaluation prompt, we were able to reduce bias and minimize hallucination in the assessment process.
}

\begin{table}[!b]
\scriptsize
\centering
\begin{tabularx}{\linewidth}{@{}lX@{}}
\toprule
\textbf{Rating} & \textbf{Description} \\
\midrule
\textbf{5 — Excellent} &
Disruption fully fixed; only necessary edits; traveler constraints and persona respected; plan coherent and realistic. \\
\textbf{4 — Good} &
Handled correctly but minor flaws like tiny unnecessary change (Flexiventurer) or very small out-of-scope tweak (Planbound). \\
\textbf{3 — Average} &
Fixed but sloppy: extra edits beyond scope (Planbound) or unnecessary changes (Flexiventurer); minor timing/sequence issues. \\
\textbf{2 — Poor} &
Superficial fix: only an acknowledgement; disrupted POI still present; major ordering/timing incoherence. \\
\textbf{1 — Very Poor} &
Disruption ignored: plan unchanged or disrupted POI untouched; edits irrelevant to the disruption. \\
\bottomrule
\end{tabularx}
\caption{Rubric for LLM judging of disrupted travel plans}
\label{tab:judge_rubric}
\end{table}


\subsection{Results}

\begin{figure}[t]
        \centering
        \includegraphics[width = 0.50\textwidth]{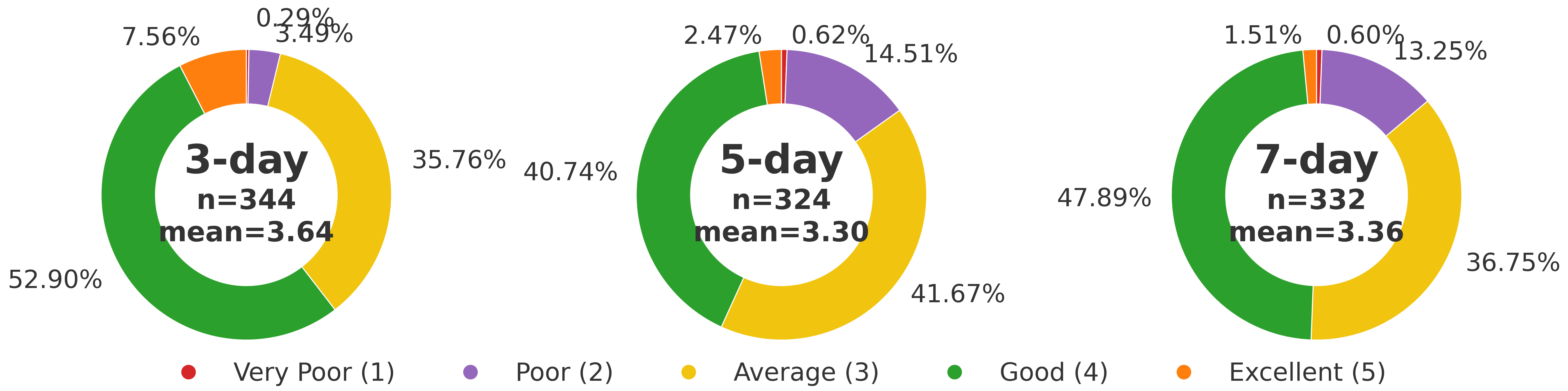}
        \caption{Llama-3.1 8B Instruct Evaluation Results}

        \label{fig:llama_judge}
    \end{figure}

\textcolor{black}{Fig.~\ref{fig:llama_judge} shows a clear pattern: 3-day plans are judged strongest overall (highest mean; majority \emph{Good} with a noticeable number of \emph{Excellent} samples). 5-day skews toward \emph{Average}/\emph{Good} with very few \emph{Excellent}, yielding the lowest mean. 7-day sits between: more \emph{Good} than 5-day, but still limited \emph{Excellent}.}

\paragraph{Relation to automated metrics.}
\textcolor{black}{Comparing results in Fig.~\ref{fig:llama_judge} with the automated metric scores in Table~\ref{tab:intent-adapt} , we see agreement on overall plan quality (few outright failures). Delivery rate is saturated ($\approx$100\%), so differences emerge in constraint adherence. 5-day plans, despite having the lowest judge mean, achieve the strongest compliance: HCPR (micro/macro) 67.38/67.59 and Final-Pass Rate 41.98\%. 3-day plans, while best judged (mean 3.64), show weaker formal validity: CPR$_{\text{macro}}$ of 29.95, HCPR$_{\text{macro}}$ of 48.84, and Final-Pass Rate of 29.37\%. The 7-day bucket is intermediate: HCPR micro/macro of 55.59/57.54 and Final-Pass Rate of 32.54\%. Taken together, the `Llama-as-a-Judge' ratings reflect perceived quality (favoring 3-day), while automated scripts surface hard feasibility (favoring 5-day), with 7-day providing a balanced middle ground.}



\section{Human Evaluation of Revised Plans}
\textcolor{black}{To complement automated scoring, we conducted a human evaluation to assess LLM efficiency within our disruption-aware framework. Three domain experts independently reviewed 100 revised itineraries (sampled across 3-, 5-, and 7-day plans). For each case, they examined the original plan, the disruption, and the revision, judging responsiveness and whether updates preserved semantic, spatial, sequential aspects.}

\textcolor{black}{The study focused on how models like GPT-4o, Qwen-2.5 7B Instruct, and Phi-4-mini Instruct demonstrate reasoning, contextual understanding, and adaptability under unexpected changes. These capabilities enable dynamic, user-aligned revisions that account for real-world uncertainties, setting a new benchmark for disruption-aware trip generation. Expert judgments form the basis of the analysis reported in this section. Detailed case studies are in Appendix~\ref{app:caseStudies}.}

\subsection{General trends}
    \noindent\textbf{Reliable disruption detection, disciplined yet non-trivial revisions.} Across all samples, the GPT-4o planner consistently identified the disrupted element (venue closure, transport capacity, infeasible timing) and initiated a corrective edit. Experts noted that, while many fixes were intentionally local (high responsiveness), a measurable subset propagated beyond the impacted slot or day, indicating a deliberate but imperfect balance between targeted repair and broader itinerary stability.
    
    \noindent\textbf{Quality changes with trip length.} In 3-day itineraries, edits remained compact with limited downstream consequences, enabling strong sequentiality and semantic preservation. 5-day plans more often incurred scope drift (additional edits not strictly required by the disruption), reflecting the added complexity of multi-day dependencies. 7-day itineraries showed the widest variance: several exemplary, end-to-end consistent repairs, but also more opportunities for spatial or temporal slippage when local changes were not fully propagated.

\setlength{\abovecaptionskip}{2pt} 



\begin{figure}[!t]
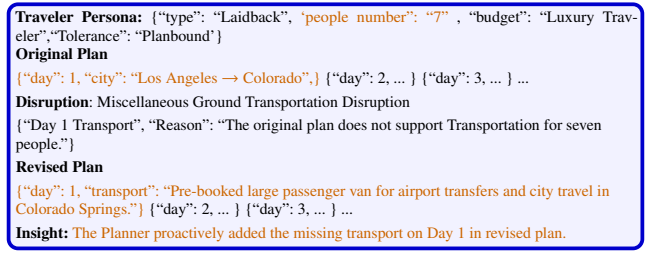

\begin{tcolorbox}[enhanced,colback=blue!5!white, colframe=blue!80!black,boxsep=0pt, left=2pt, right=2pt, top=2pt, bottom=2pt]
\scriptsize
\textbf{Traveler Persona:} \{``type'': ``Laidback'',   \textcolor{orange!80!black}{`people number'': ``7'' }, ``budget'': ``Luxury Traveler'',``Tolerance'': ``Planbound'\}

\textbf{Original Plan}

\begin{flushleft}
  \textcolor{orange!80!black}{\{``day'': 1, ``city'': ``Los Angeles $\rightarrow$ Colorado'',\}} 
 \{``day'': 2, ... \} \{``day'': 3, ... \} 
...\\
\end{flushleft}
\textbf{Disruption}: Miscellaneous Ground Transportation Disruption 
\begin{flushleft}
\{``Day 1  Transport'', ``Reason'': ``The original plan does not support Transportation for seven people.''\} \\
\end{flushleft}
\textbf{Revised Plan}

\begin{flushleft}

  \textcolor{orange!80!black}{\{``day'': 1, ``transport'': ``Pre-booked large passenger van for airport transfers and city travel in Colorado Springs.''\}} 
\{``day'': 2, ... \} \{``day'': 3, ... \} 
...\\

\end{flushleft}

\textbf{Insight:}
  \textcolor{orange!80!black}{The Planner proactively added the missing transport on Day 1 in revised plan.}
\end{tcolorbox}
\caption{Enhanced plan, handling logistics for a large group.}
\label{fig:human_evaluation_1}
\end{figure}

\begin{figure}[!t]
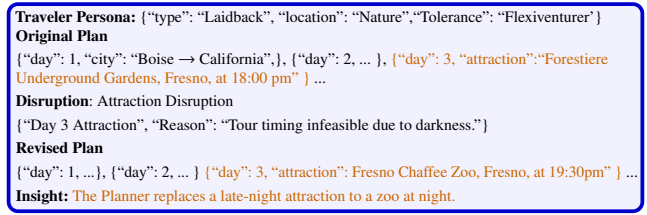

\begin{tcolorbox}[enhanced,colback=blue!5!white, colframe=blue!80!black,boxsep=0pt, left=2pt, right=2pt, top=2pt, bottom=2pt]
\scriptsize
\textbf{Traveler Persona:} \{``type'': ``Laidback'',  ``location'': ``Nature'',``Tolerance'': ``Flexiventurer'\}

\textbf{Original Plan}

\begin{flushleft}
\{``day'': 1, ``city'': ``Boise $\rightarrow$ California'',\}, \{``day'': 2, ... \}, 
\textcolor{orange!80!black}{\{``day'': 3, ``attraction'':``Forestiere Underground Gardens, Fresno, at 18:00 pm'' \}} 
...\\
\end{flushleft}
\textbf{Disruption}: Attraction Disruption 
\begin{flushleft}
\{``Day 3  Attraction'', ``Reason'': ``Tour timing infeasible due to darkness.''\} \\
 
\end{flushleft}
\textbf{Revised Plan}

\begin{flushleft}

\{``day'': 1, ...\}, \{``day'': 2, ... \} 
\textcolor{orange!80!black}{\{``day'': 3, ``attraction'': Fresno Chaffee Zoo, Fresno, at 19:30pm'' \}} 
...\\

\end{flushleft}

\textbf{Insight:}
  \textcolor{orange!80!black}{The Planner replaces a late-night attraction to a zoo at night. }

\end{tcolorbox}
\caption{Planner misses suitable alternative.}
\label{fig:human_evaluation_2}
\end{figure}

\subsection{Where the Planner did well}
   
  \noindent\textbf{Smart swaps.} When a POI was closed, mispriced, or poorly located, the system frequently selected a like-for-like alternative (similar category, nearby geography, compatible opening hours). This preserved semantic fidelity (intent and activity type), maintained spatial coherence (short transfers, clustered movement), and limited ripple effects, supporting sequentiality by keeping the rest of the day intact.

  \noindent\textbf{Human factors.} Experts observed fatigue-aware behavior: after long drives or dense activity blocks, the planner often inserted lighter follow-ups or buffer time. Such adjustments improved sequentiality (realistic pacing) and responsiveness (addressing the root strain) without diluting the day’s semantic goals.
  
  \noindent\textbf{Logistics.}   The system closed execution gaps (e.g., adding appropriate-capacity ground transport for larger parties or linking airport legs to city transfers), ensuring end-to-end feasibility. These edits improved spatial and sequential coherence while remaining minimally invasive (see Fig.~\ref{fig:human_evaluation_1}).
  
  \noindent\textbf{Persona fit.} Substitutions typically reflected stated interests and constraints (e.g., nature-aligned alternatives for ``Nature'' travelers; shopping added where requested), sustaining semantic alignment with preferences while respecting spatial and temporal realities.

\subsection{Where the Planner struggled}
    
\noindent\textbf{Superficial fixes.} A subset of cases exhibited nominal edits (e.g., renaming lodging or swapping to a venue with similar limitations) that did not materially resolve the underlying constraint. Such changes preserved surface semantics but weakened responsiveness and yielded little gain in spatial or sequential feasibility.

\noindent\textbf{Missing the root cause.} After very long transfers, the planner sometimes addressed a symptom (e.g., pushing a meal) without rebalancing the next day to account for accumulated fatigue. This under-correction harmed sequentiality (unrealistic pacing) and occasionally the day’s semantic goals, despite appearing minimal on paper.

\noindent\textbf{Real-world timing.} Sometimes, replacements overlooked operating-hour constraints, substituting a closed venue with another that was also unavailable at the proposed time. These errors broke sequentiality and spatial realism, despite intent to preserve activity type. See Fig.~\ref{fig:human_evaluation_2} for a timing miss where the alternative remained infeasible.

\noindent\textbf{Ripple effects.} Local adjustments (e.g., ``leave earlier'', ``add generic rest stops'') were not always propagated to dependent activities, leaving downstream segments compressed or misordered. Such partial propagation reduced sequentiality and sometimes undermined responsiveness, even when the initial substitution appeared reasonable.

\subsection{Analysis across Disruption Types}

\noindent\textbf{Accommodation Disruptions.}  
A consistent mitigation pattern was observed for accommodation-related disruptions. In four out of five instances, the unsuitable accommodation was directly replaced with a more appropriate alternative, tailored to traveler preferences such as size, comfort, or luxury level. In one case, instead of replacing the accommodation, the subsequent day's itinerary was adapted to accommodate traveler fatigue from a late-night check-in. These strategies reflect a high degree of responsiveness, as the system proactively addressed the disruption. However, slight reductions in the sequentiality score were observed when overnight adjustments resulted in deviations from the following day's schedule.

\noindent\textbf{Transportation Disruptions.}  
Transportation issues were typically resolved with minimal disruption to the overall itinerary structure. Adjustments included modifying departure times, inserting rest breaks, or switching to more appropriate transportation modes (e.g., from public to self-driven). These interventions preserved both the feasibility and logical flow of travel, contributing to consistently high semantic alignment and spatial coherence. While such decisions showcased a strong contextual understanding, the model occasionally opted for conservative changes, resulting in slightly lower responsiveness in cases of minor disruptions.

\noindent\textbf{Attraction Disruptions.}  
Mitigation strategies for disrupted attractions followed a two-fold approach: replacing the activity or rescheduling it to a more suitable time. For example, closed or contextually inappropriate attractions (e.g., hiking in low light) were either swapped for alternatives or moved to earlier in the day. These adaptations maintained both user's intent and temporal structure, yielding high semantic and sequentiality scores. Rarely,  spatial score declined marginally when replacements introduced longer travel distances.

\noindent\textbf{Miscellaneous Disruptions.}  
A broad range of mitigation responses were observed for disruptions related to restaurants, fatigue, and profile mismatches. For example, closed dining venues were substituted with similar alternatives, while travelers experiencing fatigue were assigned more relaxed schedules, including later starts and reduced physical demands. Adjustments were also made to better align with traveler profiles, for instance, replacing luxury accommodations for budget travelers, and vice versa. These nuanced modifications resulted in elevated semantic scores, reflecting the model’s sensitivity to user-specific needs. However, such refinements occasionally came at the expense of spatial efficiency, particularly when suitable alternatives were not in close proximity.

\begin{figure}[!t]
\begin{tcolorbox}[enhanced,colback=blue!5!white, colframe=blue!80!black,boxsep=0pt, left=2pt, right=2pt, top=2pt, bottom=2pt]
\scriptsize
\textbf{Traveler Persona:} \{``type'': ``Laidback'', ``purpose'': ``Nature'',``cuisine'':``Mexican, Italian'', ``budget'': ``Economical'', ``pref'': ``Beaches'',``Tolerance'': ``Flexi-Venturer''\}

\textbf{Original Plan}

\begin{flushleft}
\{``day'': 1, ... \} \{``day'': 2, ... \} \{``day'': 3, ... \} \{``day'': 4, ``city'': ``Durango'', ``transport'': ``-'',, ``breakfast'': ``-'', ``attraction'': ``-'', ``lunch'': ``\textcolor{orange!80!black}{Zia Taqueria, Durango}'', ``dinner'': ``-'', ``stay'': ``Modern Victorian''\} \\
\end{flushleft}
\textbf{Disruption}
\begin{flushleft}
\{``Day 4  Lunch'': ``Zia Taqueria, Durango'',
``Disruption Category':`` Restaurants '', 
``Reason'':``The Zia taquerias Tuesday deals attract college students, leading to long waits and limited seating for groups.''\}   
\end{flushleft}
\textbf{Revised Plan}

\begin{flushleft}
\{``day'': 1, ... \} \{``day'': 2, ... \} \{``day'': 3, ... \} \{``day'': 4, ``city'': ``Durango'', ``transport'': --, ``breakfast'': ``Durango Doughworks'', ``attraction'': ``Garden of the Gods'', ``lunch'': ``\textcolor{orange!80!black}{Michel's Corner Crepes, Durango}'', ``dinner'': ``Mountain Shadows'', ``stay'': ``Modern Victorian''\} \\
\end{flushleft}
\textbf{Semantic Issue:}
  \textcolor{orange!80!black}{The Restaurant Disruption in Day 4 is handled by the LLM generated plan, but it fails to capture the cuisine preference of the traveler that is Mexican or Italian, rather it replaces a Mexican style restaurant by a French type restaurant.}
\end{tcolorbox}
\caption{Semantic issues for a 5-day plan}
\label{fig:cs1}
\end{figure}

\begin{figure}
\begin{tcolorbox}[colback=blue!5!white, colframe=blue!80!black, boxsep=0pt, left=2pt, right=2pt, top=2pt, bottom=2pt]
\scriptsize
\textbf{Traveler Persona:} \{``type'': ``Laidback'', ``purpose'': ``History'', ``budget'': ``Economical'', ``pref'': ``Museums'',``Tolerance'': ``Plan-Bound''\}

\textbf{Initial Plan }

\begin{flushleft}
\{``day'': 1, ``city'': ``San Diago'', ``poi\_sequence'': [``Calibro National Monument'', ``Coronado Bridge' ...]\} \\
\{``day'': 2, ``city'': ``San Diago'', ``poi\_sequence'': [``\textcolor{red}{Seaport Village (377.43m away)}'', ``Petco Park'' ...]\} \\
\{``day'': 3, ``city'': ``San Diago'', `poi\_sequence'': [``Return to Seattle by \textcolor{red}{flight number F0293908,[11:35, 13:05]'']\}} \\
\end{flushleft}
\textbf{Disruption}
\begin{flushleft}
    ``Day 2 Attraction'': ``Seaport Village'', 
    ``Disruption Category'':``Attraction'', 
    ``Reason'': ``There is an ongoing plan called Seaport San Diego to redevelop Seaport Village, because of which tourists are not allowed.''\\
\end{flushleft} 
\textbf{Revised Plan }
\begin{flushleft}
\{``day'': 1, ``city'': ``San Diago'', ``poi\_sequence'': [``Calibro National Monument'', ``Coronado Bridge' ...]\} \\
\{``day'': 2, ``city'': ``San Diago'', ``poi\_sequence'': [\textcolor{red}{``La Jolla Shores Park (243.08m away)''}, ``Petco Park'' ...]\} \\
\{``day'': 3, ``city'': ``San Diago'', `poi\_sequence'': [``Return to Seattle by \textcolor{red}{flight number F0244163 [12:37, 13:52]''} ]\} \\
\end{flushleft}
\textbf{Spatial Divergence} \textcolor{red}{Even with the attraction swap and flight changes the distances of POIs to nearest transit stayed within similar ranges (50-200m), thus preserving the spatial divergence $A_{spa}$.}
\end{tcolorbox}
\caption{Preservation of spatial convenience for a 3 day plan}
\label{fig:cs2}
\end{figure}

\noindent\textbf{Summary.}  
Overall, the analysis demonstrates that the model performs well across most disruption types, exhibiting particularly strong performance in responsiveness and semantic preservation. Minor trade-offs in sequentiality and spatial planning were observed in specific scenarios, often arising from the inherent complexity of balancing user intent with contextual feasibility. These findings highlight the model's ability to effectively adapt plans in a user-aligned and disruption-aware manner.

\noindent\textbf{Analysis of Examples.} 
The two snippets in Figs.~\ref{fig:cs1} and~\ref{fig:cs2} show the robust performance of GPT-4o for a 5-day and a 3-day case from our dataset. The model makes multiple semantic mistakes in 5-day while maintaining the spatial coherence for 3-day. This snippet indicates substantial scope for further improvement in this area.

\section{Conclusion} 
We introduce \system{}, the benchmark to evaluate LLMs' ability to prescribe travel itineraries, in the face of real-world disruptions. To the best of our knowledge, this is the first work to integrate disruption severity levels with user-specific tolerance profiles in the context of travel planning. By jointly modeling the granularity of disruptions (\textit{step-level}, \textit{day-level}, and \textit{plan-level}), alongside traveler flexibility types (e.g., \textit{Flexi-Venturer} vs. \textit{Plan-Bound}), our framework captures a significantly broader and more realistic spectrum of traveler behaviors. Moreover, \textcolor{black}{alongside LLM-act-as-a Judge and manual evaluation by human experts}, we introduce a set of novel evaluation metrics to evaluate LLM-generated plans, via measuring preservation of user intent, the responsiveness to disruptions, and the adaptability of the revised plans. This nuanced integration facilitates personalized and context-aware itinerary adaptations, effectively bridging a critical gap in prior work, which does not care for disruptions.


\bibliographystyle{ACM-Reference-Format}
\bibliography{reference}

\appendix


\vspace{0.5cm}


\noindent {\Large \textbf{Overview of Appendix Sections}}

\begin{itemize}
  
  \item Appendix~\ref{app:hyperparams}: Hyper-parameter Details for Reproducibility
  \item Appendix~\ref{app:humanAnnotation}: Human Annotation Guidelines
  \item Appendix~\ref{app:resultsAnalysis}: Detailed Analysis of Results
  \item Appendix~\ref{app:caseStudies}: Case Studies
\end{itemize}

\section{Hyper-parameter Details for Reproducibility}
\label{app:hyperparams}
In this paper, we have used GPT-4o, Qwen2.5-7B-Instruct, and Phi4-mini-Instruct LLM models to generate the revised travel plan based on our dataset. The hyperparameter details of the three models are listed below:\\

\begin{table}[h]
\centering
\scriptsize
\begin{tabular}{llp{0.25\columnwidth}p{0.25\columnwidth}}
\toprule
\textbf{Parameter} & \textbf{GPT-4o} & \textbf{Qwen2.5-7B instruct} & \textbf{Phi-4- mini instruct} \\
\midrule
Model Source      & OpenAI  & Alibaba (via HuggingFace) & Microsoft (via HuggingFace) \\
Model Size        & Unknown  & 7B  & 2.7B \\
Temperature       & 0.0 & 0.0   & 0.0 \\
Top-p             & 1.0 &1.0 &1.0\\
Max Tokens        & 10,000 & 3072  & 4096 \\ \\
Frequency Penalty & 0.0 & 0.0 & 0.0 \\ \\
Presence Penalty  & 0.0 & 0.0 & 0.0 \\
System Prompt     & Github Repo  & Github Repo  & Github Repo \\
Version           & May 2024 & Sep 2024  & Feb 2025 \\
\bottomrule
\end{tabular}
\caption{Model Inference Hyperparameters}
\label{tab:model-info}
\end{table}

\begin{table}
    \centering
    \scriptsize
    \begin{tabular}{|c|p{0.9\columnwidth}|}
        \hline
        \textbf{\#} & \textbf{Annotation Guidelines} \\
        \hline
        1 & The objective is to generate an alternative feasible plan following the occurrence of a disruption.\\
        \hline
        2.& Input: Current plan, disruption, disruption severity, disruption\_tolerance\\ &Output: alternative feasible plan.\\
        \hline
        3 & The alternative plan should introduce only the minimal necessary modifications, and the updated list of Points of Interest (POIs) must not include any fabricated or unverified information. The feasibility will be checked by us using scripts. \\
        \hline
        4 & The annotators have to keep in mind that the disruption mitigation process depends on two factors: both Disruption\_severity and Disruption\_tolerance.Refer Table~\ref{tab:disruption_types}. \\
        \hline
        5 & The extent of changes should correspond to the disruption\_severity: \\ &\textbf{Step-level:} disruptions should involve the least amount of change, changing only the disrupted event.\\ &\textbf{Day-level:} disruptions may require moderate adjustment, meaning modifying the disrupted day’s itinerary only.\\ &\textbf{Plan-level:} disruptions may necessitate more extensive revisions. The annotators are free to change the entire plan, if necessary.\\
        \hline
        6 & Annotators must consider the traveler’s disruption\_tolerance attribute (Flexiventurer / Planbound) when modifying the plan, and should provide clear remarks explaining the adjustments made.  \\
        \hline
        7 & The travelers are categorized based on their “disruption\_tolerance”:\\ &\textbf{Flexiventurer:} Open to last-minute modifications, such as rearranging the itinerary by substituting POIs from subsequent days to optimize the overall experience.\\ &\textbf{Planbound:} Prefer to adhere strictly to the original itinerary and are reluctant to accept any significant changes. \\
        \hline
        8 & The annotation process must consider the traveler's classification. If the traveler is identified as “Planbound”,  the scope of revision must strictly correspond to the disruption\_severity. Specifically, for step-level disruptions, only the affected event should be modified; for day-level disruptions, modifications must be limited to the POIs scheduled for that particular day; and for plan-level disruptions, broader itinerary changes are permitted.\\ & In contrast, for “Flexiventurer” travelers, there is no constraint linking the revision scope to the disruption severity. Annotators are granted the flexibility to revise the itinerary as deemed appropriate. However, any such modifications must be accompanied by a clear justification in the “Remarks” column."\\
        \hline
        9 & While generating the revised POI list, annotators must preserve the original user persona characteristics. For example: A laidback traveler may prefer a schedule with 1-2 attractions per day, even when additional options are available. An economical traveler would prioritize budget-friendly options over more expensive alternatives. \\
        \hline
        10 &Annotators are required to provide truthful and contextually appropriate responses by generating a revised travel plan that differs from the original itinerary, effectively addressing and mitigating the impact of the identified disruption\\
        \hline
        11.&Annotators should apply their best judgment to ensure that the updated plans are both practical and realistic. Any major decision-making rationale must be explicitly documented in the Remarks section.\\
        \hline
    \end{tabular}
    \caption{Guidelines for Annotation of Revised Travel Plans and Remarks}
    \label{tab:annotation_guidelines}
\end{table}

\paragraph{Server Configuration Details.}
\label{appendix:system-config}

All local inference experiments were conducted on a high-performance server with the following system specifications:

\begin{itemize}
    \item \textbf{Operating System:} Ubuntu 24.04.2 LTS
    \item \textbf{CPU:} Dual AMD EPYC 9474F (2 × 48-Core, 192 threads total)
    \item \textbf{GPU:} NVIDIA L40, 46 GB VRAM
    \item \textbf{CUDA Version:} 12.9
    \item \textbf{NVIDIA Driver Version:} 575.64.03
    \item \textbf{Python Version:} 3.10
    \item \textbf{Inference Libraries:}
    \begin{itemize}
        \item \texttt{transformers==4.53.2}
        \item \texttt{torch==2.5.1}
        \item \texttt{accelerate==1.9.0}
    \end{itemize}
    \item \textbf{Memory:} 46 GB GPU memory, 1TiB system RAM
    \item \textbf{Model Hosting:}
    \begin{itemize}
        \item GPT-4o: Accessed via OpenAI API
        \item Qwen2.5-7B-Instruct: Inferred locally using Hugging Face model \texttt{Qwen/Qwen2.5-7B-Instruct}
        \item Phi-4-mini-Instruct: Inferred locally using Hugging Face model \newline \texttt{microsoft/Phi-4-mini-instruct}
    \end{itemize}
\end{itemize}


\section{Human Annotation Guidelines}
\label{app:humanAnnotation}

\paragraph{Guidelines for Annotation}
Annotators were instructed to manually identify potential day-wise disruptions for each travel plan. The goal was to sample disruptions in a balanced manner across key categories: \textit{Accommodation}, \textit{Attractions}, \textit{Restaurants}, \textit{Transport}, and \textit{Miscellaneous}. Once disruptions were sampled, annotators were required to generate a corresponding Revised Plan. This plan had to be created using the \textit{Disruption Query}, the \textit{Initial Plan}, and the \textit{Reference Information}. The annotated plan was expected to strictly adhere to the specified \textit{User Persona}, \textit{Disruption Severity}, and the \textit{Traveler’s Tolerance Level}.
Please refer to Table~\ref{tab:annotation_guidelines} for detailed annotation guidelines.

\paragraph{Annotator Demographics}
The demographic distribution of annotators reflects a diverse range of educational backgrounds and experience levels. English proficiency distribution of our graduate student annotators is as follows: 8 years (6.7\%), 12 years (26.7\%), 16 years (53.3\%), 20 years (13.3\%). 
This shows that the majority of annotators have received 12 to 20 years of formal English education, indicating a high level of language proficiency. Age distribution of our graduate student annotators is as follows: 18 years (3.3\%), 19 years (20.0\%), 20 years (26.7\%), 21 years (33.3\%), 22 years (6.7\%), 23 years (10.0\%).
Thus, the age distribution
is centered around 20 to 25 years, suggesting that most annotators are young graduate students in early-to-mid adulthood. Gender representation (Male: 66.7\%, Female: 33.3\%)
 also shows a balanced participation across genders among the 30 graduate student annotators. These demographics suggest that our annotators possess strong language and reasoning abilities, along with the cognitive maturity required to effectively evaluate and annotate revised travel plans.

\section{Detailed Analysis of Results}
\label{app:resultsAnalysis}

In this section we analyze how the models perform based on the disruption severity (Step, Day or Plan) irrespective of travel duration. Results are presented in Tables~\ref{tab:preservation-intent-severity} and~\ref{tab:adaptability-severity-level}.

\paragraph{Preservation of Intent:} Table~\ref{tab:preservation-intent-severity} shows that GPT-4o maintains a perfect delivery rate across most disruption levels, with strong performance in both commonsense (CPR) and hard constraint (HCPR) pass rates, particularly excelling at Step and Plan-level disruptions. Qwen-2.5 7B instruct also sustains a high delivery rate, but its HCPR scores remain notably lower, especially under Step-level disruptions, indicating weaker handling of strict constraints. While the models show stable CPR performance, GPT-4o consistently achieves better constraint satisfaction overall indicating the model's strength to capture the intent across all the disruption severity levels.

\paragraph{Semantic Adaptability ($A_{\text{sem}}$):} Table~\ref{tab:adaptability-severity-level} reflects that GPT-4o achieves the lowest semantic score at Step-level (0.05), indicating strong preservation of original plan intent under fine-grained disruptions. However, it worsens slightly at Day (0.41) and Plan-level (0.17). In contrast, Qwen2.5-7B-Instruct maintains slightly higher and relatively constant semantic scores (0.40-1.46), suggesting that its plan revisions diverge more from the original semantic context, particularly under Day-level disruptions.

\paragraph{Spatial Adaptability ($A_{\text{spa}}$):} GPT-4o exhibits stable spatial consistency across all levels, with its best score at Day-level (0.21). Qwen2.5-7B-Instruct instruct struggles at Step-level with a high score of 5.49, implying poor location coherence, but performs best at Day-level (0.01). This reflects the fact that Qwen2.5-7B-Instruct adapts better when the disruption granularity is coarser.

\paragraph{Sequential Adaptability ($A_{\text{seq}}$):} GPT-4o shows excellent temporal consistency at Day and Plan-levels (as low as 5), but much higher at Step-level (82.36), indicating difficulty in preserving activity sequences during fine-grained changes. On the other hand, Qwen2.5-7B-Instruct performs moderately with a score of nearly 30, suggesting GPT-4o performs better when maintaining the sequentiality of the POI list with respect to the original plan for day-level and plan-level disruptions.

\paragraph{Responsiveness rate:} Qwen2.5-7B-Instruct outperforms GPT-4o while checking the mitigation rate of the models between the final revised and initial plan. We can observe from Table~\ref{tab:adaptability-severity-level} that for Qwen2.5-7B-Instruct the rate is nearly 100\% while GPT-4o performance degrades as the severity level increases.

\paragraph{Summary.} In this paper we analyzed capacity of GPT-4o,  Qwen2.5-7B-Instruct to handle the disruptions for both when the dataset is divided among Travel Durations (3, 5, 7-days) and Disruption Severity (Step, Day, and Plan Levels). From Table~\ref{tab:preservation-intent-severity} and Table~\ref{tab:adaptability-severity-level}, we can infer that, in the day-wise setting, GPT-4o maintains high sequential consistency and improving spatial scores with longer plans, showing stable adaptability. Qwen2.5-7B-Instruct and Phi-4-mini-Instruct achieve higher responsiveness but struggle with semantic and spatial coherence, especially for 3-day plans. In the severity-wise setting, GPT-4o excels in semantic alignment under step-level disruptions but drops in responsiveness. Meanwhile, Qwen2.5-7B-Instruct retain high responsiveness across severities but shows poor sequential alignment at finer-grained disruptions.

\begin{table}[!t]
\centering
\scriptsize
\setlength{\tabcolsep}{3pt}

\begin{tabular}
{p{0.12\columnwidth}p{0.12\columnwidth}p{0.11\columnwidth}p{0.09\columnwidth}p{0.09\columnwidth}p{0.09\columnwidth}p{0.09\columnwidth}p{0.11\columnwidth}}
\toprule
\multirow{2}{0.12\columnwidth}{\textbf{Model Name}} & 
\multirow{2}{0.12\columnwidth}{\textbf{Plan\newline Duration}} & 
\multirow{2}{0.11\columnwidth}{\textbf{Delivery Rate}$\uparrow$} & 
\multicolumn{2}{c}{\textbf{CPR}$\uparrow$} & 
\multicolumn{2}{c}{\textbf{HCPR}$\uparrow$} & 
\multirow{2}{0.11\columnwidth}{\textbf{Final Pass Rate}$\uparrow$} \\ 
& & & \textbf{Micro} & \textbf{Macro} & \textbf{Micro} & \textbf{Macro} & \\
\midrule
\midrule
\multirow{3}{0.12\columnwidth}{GPT-4o} 
& Step-Level & 100.00 & \textbf{90.29} & 32.46 & 47.68 & 46.56 & 31.15  \\
& Day- level & 100.00 & 89.94 & \textbf{33.78} & 54.15 & \textbf{59.68} & \textbf{33.45}   \\
& Plan-level & 78.69 & 70.49  & 30.17 & \textbf{56.09} & 55.74 & 29.51 \\
\midrule
\multirow{3}{0.12\columnwidth}{Qwen2.5-7B-Instruct} 
& Step Level & 99.65  & 86.98  & 24.49 & 42.47 & 44.22 & 24.49  \\
& Day- level &  99.33 & 86.41  & 25.43 & 44.23 & 45.43 & 24.07   \\
& Plan-level &  98.71 & 85.86  & 28.14 & 51.54 & 49.36 & 28.14 \\
\bottomrule
\end{tabular}
\caption{Preservation of user intent scores for different Models based on Disruption Severity level.}
\label{tab:preservation-intent-severity}
\end{table}

\begin{table}[!t]
\centering
\scriptsize
\tabcolsep3pt
\begin{tabular}{p{0.14\columnwidth}p{0.13\columnwidth}p{0.14\columnwidth}p{0.14\columnwidth}p{0.15\columnwidth}p{0.15\columnwidth}}
\toprule
\textbf{Model Name} & 
\textbf{Disruption\newline Severity} & 
\textbf{Semantic Score \newline($A_{\text{sem}}$)} $\downarrow$ & 
\textbf{Spatial Score \newline($A_{\text{spa}}$)} $\downarrow$& 
\textbf{Sequential Score \newline($A_{\text{seq}}$)} $\downarrow$& 
\textbf{Responsive-ness Rate} $\uparrow$\\
\midrule
\multirow{3}{0.16\columnwidth}{GPT-4o} 
& Step-level & \textbf{0.05} & 0.65 & 82.36 & 94.11\\
& Day-level & 0.41 & 0.21 & 4.98 & 78.84 \\
& Plan-level & 0.17 & 1.41 & \textbf{4.55} & 76.76 \\
\midrule
\multirow{3}{0.16\columnwidth}{Qwen2.5-7B-instruct} 
& Step-level & 0.40 & 5.49 & 34.06 & 96.39\\
& Day-level & 1.46 & \textbf{0.01} & 32.54 & \textbf{99.67} \\
& Plan-level & 0.41 & 1.31 & 31.42 & 99.57 \\
\bottomrule
\end{tabular}
\caption{Adaptability ($A_{\text{sem}}, A_{\text{spa}}, A_{\text{seq}}$) and Responsiveness scores (\%) for different models based on Disruption Severity level.}
\label{tab:adaptability-severity-level}
\end{table}

\newpage
\section{Case Studies}
\label{app:caseStudies}
In this section we showcase a few of the examples that highlight the model's strengths and weaknesses while handling the disruptions.
\subsection{Example showing how GPT-4o effectively mitigated the disruption while maintaining the Semantic and Traveler tolerance level.}

\begin{center}
\begin{tcolorbox}[colback=green!5!white, colframe=blue!80!black, arc=1mm, boxrule=0.5pt, width=\columnwidth]
\scriptsize
\textbf{Initial Plan}
\{``idx'': 27, ``row\_number'': ``532-1'', ``JSON'': \{``org'': ``Miami'', ``dest'': ``Baltimore'', ``days'': 3, ``visiting\_city\_number'': 1, ``date'': [``2024-11-03'', ``2024-11-04'', ``2024-11-05''], ``people\_number'': 2, ``local\_constraint'': \{``house rule'': null, ``cuisine'': null, ``room type'': null, ``transportation'': null, ``event'': [``Arts \& Theatre''], ``attraction'': null\}, ``budget'': 1800.0, ``query'': null, ``level'': ``medium''\}, ``persona'': ``Traveler Type: Adventure Seeker; Purpose of Travel: Adventure; Spending Preference: Luxury Traveler; Location Preference: Mountains'', ``disruption tolerance'': ``Planbound'', ``plan'': [\{``days'': 1, ``current\_city'': ``from Miami to Baltimore'', ... \},\\
\{``days'': 2, ``current\_city'': ``Baltimore'', ... \},\\
\{``days'': 3, ``current\_city'': ``from Baltimore to Miami'', ``transportation'': ``Flight Number: F2123935, from Baltimore to Miami, Departure Time: 12:11, Arrival Time: 14:31'', ``breakfast'': ``Blue Moon Cafe, Baltimore'', ``attraction'': ``Fort McHenry National Monument And Historic Shrine, Baltimore'', ``lunch'': ``-'', ``dinner'': ``-'', ``accommodation'': ``-'', ``event'': ``-'', ``point\_of\_interest\_list'': ``Luxe Houseboat: Stunning Inner Harbor 360 Views, stay from 07:00 to 07:30, nearest transit: SHOT TOWER METRO West Entrance, 841.96m away;Blue Moon Cafe, breakfast from 08:00 to 09:00, nearest transit: SHOT TOWER STATION (METRO) sb, 1161.67m away;Fort McHenry National Monument And Historic Shrine, visit from 09:30 to 11:30, nearest transit: SHOT TOWER STATION (METRO) sb, 3547.78m away.''\}]\}

\textbf{Disruption Info: }
Day 3  Attraction: Fort McHenry National Monument  \\
Disruption Category: Attractions  \\
Reason: Historic sites often have limited morning hours, and visit ends very close to flight check-in.\\
Disruption Detection Timestamp: 1 day before \\
Disruption Severity: Day-level  \\
Confidence Level: Medium \\

\textbf{Revised Plan}
\{``acknowledgement'': ``Attractions Disruption identified and understood. Proceeding with appropriate revisions based on disruption severity and traveler tolerance.'', ``idx'': 27, ``row\_number'': ``532-1'', ``JSON'': \{``org'': ``Miami'', ``dest'': ``Baltimore'', ``days'': 3, ``visiting\_city\_number'': 1, ``date'': [``2024-11-03'', ``2024-11-04'', ``2024-11-05''], ``people\_number'': 2, ``local\_constraint'': \{``house rule'': null, ``cuisine'': null, ``room type'': null, ``transportation'': null, ``event'': [``Arts \& Theatre''], ``attraction'': null\}, ``budget'': 1800.0, ``query'': null, ``level'': ``medium''\}, ``persona'': ``Traveler Type: Adventure Seeker; Purpose of Travel: Adventure; Spending Preference: Luxury Traveler; Location Preference: Mountains'', ``disruption tolerance'': ``Planbound'', ``plan'': [\{``days'': 1, ``current\_city'': ``from Miami to Baltimore'', ...\},\\

\{``days'': 2, ``current\_city'': ``Baltimore'', ... \},\\
\{``days'': 3, ``current\_city'': ``from Baltimore to Miami'', ``transportation'': ``Flight Number: F2123935, from Baltimore to Miami, Departure Time: 12:11, Arrival Time: 14:31'', ``breakfast'': ``Blue Moon Cafe, Baltimore'', ``attraction'': ``Inner Harbor, Baltimore'', ``lunch'': ``-'', ``dinner'': ``-'', ``accommodation'': ``-'', ``event'': ``-'', ``point\_of\_interest\_list'': ``Luxe Houseboat: Stunning Inner Harbor 360 Views, stay from 07:00 to 07:30, nearest transit: SHOT TOWER METRO West Entrance, 841.96m away;Blue Moon Cafe, breakfast from 08:00 to 09:00, nearest transit: SHOT TOWER STATION (METRO) sb, 1161.67m away;Inner Harbor, visit from 09:30 to 10:30, nearest transit: SHOT TOWER STATION (METRO) sb, 3547.78m away.''\}]\}\\

\end{tcolorbox}
\end{center}
\textbf{Analysis:} The model mitigated the Day 3 disruption which is a \textit{Day-level disruption} with another apprpriate attraction type. The model was also able to capture the tolerance level of traveler (\textit{Plan-bound)} as all other day's POI list remains same.

\subsection{Example citing the GPT-4o model's struggles to mitigate the Disruption for 5-day travel Planning}
\begin{center}

\begin{tcolorbox}[enhanced, breakable,colback=green!5!white, colframe=blue!80!black, arc=1mm, boxrule=0.5pt, width=\columnwidth]
\scriptsize
\textbf{Initial Plan}
\{``idx'': 177, ``row\_number'': ``65-1'', ``JSON'': \{``org'': ``Lynchburg'', ``dest'': ``North Carolina'', ``days'': 5, ``visiting\_city\_number'': 2, ``date'': [``2024-11-02'', ``2024-11-03'', ``2024-11-04'', ``2024-11-05'', ``2024-11-06''], ``people\_number'': 1, ``local\_constraint'': \{``house rule'': ``parties'', ``cuisine'': null, ``room type'': ``entire home'', ``transportation'': null, ``event'': null, ``attraction'': ``Nature \& Parks''\}, ``budget'': 1800.0, ``query'': null, ``level'': ``hard''\}, ``persona'': ``Traveler Type: Laidback Traveler; Purpose of Travel: Nature; Spending Preference: Economical Traveler; Location Preference: Beaches'', ``disruption tolerance'': ``Flexiventurer'', ``plan'': [\{``days'': 1, ``current\_city'': ``from Lynchburg to Greensboro'', ...\} \\
\{``days'': 2, ``current\_city'': ``Greensboro'', ... \}, ...\\

\{``days'': 4, ``current\_city'': ``Charlotte'', ``transportation'': ``-'', ``breakfast'': ``The Fig Tree Restaurant, Charlotte'', ``attraction'': ``Mint Museum Uptown, Charlotte; The Bank Of America Stadium, Charlotte'', ``lunch'': ``Princess Charlotte, Charlotte'', ``dinner'': ``Cabo Fish Taco Baja Seagrill, Charlotte'', ``accommodation'': ``Tippah Treehouse Retreat, Charlotte'', ``event'': ``-'', ``point\_of\_interest\_list'': `` Tippah Treehouse Retreat, stay from 09:00 to 09:30, nearest transit: The Plaza \& Marguerite Ave, 381.54m away; The Fig Tree Restaurant, visit from 10:00 to 10:30, nearest transit: 7th St \& Armory Dr, 103.61m away; Mint Museum Uptown, visit from 11:30 to 14:00, nearest transit: Tryon St \& Stonewall St, 100.42m away; Princess Charlotte, visit from 14:30 to 15:30, nearest transit: Beaty St @ 866, 6465940.22m away; The Bank Of America Stadium, visit from 16:00 to 18:30, nearest transit: Stonewall St \& Mint St, 270.52m away; Cabo Fish Taco Baja Seagrill, visit from 19:30 to 21:30, nearest transit: Davidson \& 35th, 57.25m away; Tippah Treehouse Retreat, stay from 22:00 to 09:00, nearest transit: The Plaza \& Marguerite Ave, 381.54m away.''\},\\
\{``days'': 5, ``current\_city'': ``from Charlotte to Lynchburg'',
... \}]\}

\textbf{Disruption Info: }\\
Day 4  Attraction: The Bank of America Stadium\\
Disruption Category: Attractions\\
Reason: Stadium tours may be unavailable due to game-day prep or private events.\\
Disruption Detection Timestamp: 1 day before (via official stadium tour site)\\
Disruption Severity: Step-level\\
Confidence Level: Medium

\textbf{Revised Plan}
\{``acknowledgement'': ``Attractions Disruption identified and understood. Proceeding with appropriate revisions based on disruption severity and traveler tolerance.'', ``idx'': 177, ``row\_number'': ``65-1'', ``JSON'': \{``org'': ``Lynchburg'', ``dest'': ``North Carolina'', ``days'': 5, ``visiting\_city\_number'': 2, ``date'': [``2024-11-02'', ``2024-11-03'', ``2024-11-04'', ``2024-11-05'', ``2024-11-06''], ``people\_number'': 1, ``local\_constraint'': \{``house rule'': ``parties'', ``cuisine'': null, ``room type'': ``entire home'', ``transportation'': null, ``event'': null, ``attraction'': ``Nature \& Parks''\}, ``budget'': 1800.0, ``query'': null, ``level'': ``hard''\}, ``persona'': ``Traveler Type: Laidback Traveler; Purpose of Travel: Nature; Spending Preference: Economical Traveler; Location Preference: Beaches'', ``disruption tolerance'': ``Flexiventurer'', ``plan'': [\{``days'': 1, ``current\_city'': ``from Lynchburg to Greensboro'', 
...\}\\

\{``acknowledgement'': ``Attractions Disruption identified and understood. Proceeding with appropriate revisions based on disruption severity and traveler tolerance.'', ``idx'': 27, ``row\_number'': ``532-1'', ``JSON'': \{``org'': ``Miami'', ``dest'': ``Baltimore'', ``days'': 3, ``visiting\_city\_number'': 1, ``date'': [``2024-11-03'', ``2024-11-04'', ``2024-11-05''], ``people\_number'': 2, ``local\_constraint'': \{``house rule'': null, ``cuisine'': null, ``room type'': null, ``transportation'': null, ``event'': [``Arts \& Theatre''], ``attraction'': null\}, ``budget'': 1800.0, ``query'': null, ``level'': ``medium''\}, ``persona'': ``Traveler Type: Adventure Seeker; Purpose of Travel: Adventure; Spending Preference: Luxury Traveler; Location Preference: Mountains'', ``disruption tolerance'': ``Planbound'', ``plan'': [\{``days'': 1, ``current\_city'': ``from Miami to Baltimore'', ... \},\\
\{``days'': 2, ``current\_city'': ``Baltimore'', ... \},\\
\{``days'': 4, ``current\_city'': ``Charlotte'', ``transportation'': ``-'', ``breakfast'': ``The Fig Tree Restaurant, Charlotte'', ``attraction'': ``Mint Museum Uptown, Charlotte; The Bank Of America Stadium, Charlotte'', ``lunch'': ``Princess Charlotte, Charlotte'', ``dinner'': ``Cabo Fish Taco Baja Seagrill, Charlotte'', ``accommodation'': ``Tippah Treehouse Retreat, Charlotte'', ``event'': ``-'', ``point\_of\_interest\_list'': `` Tippah Treehouse Retreat, stay from 09:00 to 09:30, nearest transit: The Plaza \& Marguerite Ave, 381.54m away; The Fig Tree Restaurant, visit from 10:00 to 10:30, nearest transit: 7th St \& Armory Dr, 103.61m away; Mint Museum Uptown, visit from 11:30 to 14:00, nearest transit: Tryon St \& Stonewall St, 100.42m away; Princess Charlotte, visit from 14:30 to 15:30, nearest transit: Beaty St @ 866, 6465940.22m away; The Bank Of America Stadium, visit from 16:00 to 18:30, nearest transit: Stonewall St \& Mint St, 270.52m away; Cabo Fish Taco Baja Seagrill, visit from 19:30 to 21:30, nearest transit: Davidson \& 35th, 57.25m away; Tippah Treehouse Retreat, stay from 22:00 to 09:00, nearest transit: The Plaza\& Marguerite Ave, 381.54m away.''\},\\
\{``days'': 5, ``current\_city'': ``from Charlotte to Lynchburg'', 
... \}]\}
\end{tcolorbox}
\end{center}

\textbf{Analysis:} Even though the model acknowledges the Day 4 Disruption, but while generating the revised plan it fails to mitigate the Disrupted Attraction from POI list. This shows the declining Responsiveness rate when the Travel Duration increases.\\

\subsection{Example showing the GPT-4o's model struggles to capture the Disruption Severity and Traveler Tolerance Level}
\begin{center}

\begin{tcolorbox}[enhanced, breakable, colback=green!5!white, colframe=blue!80!black, arc=1mm, boxrule=0.5pt, width=\columnwidth]
\scriptsize
\textbf{Initial Plan}
\{``idx'': 37, ``row\_number'': 401, ``JSON'': \{``org'': ``Miami'', ``dest'': ``Louisville'', ``days'': 3, ``visiting\_city\_number'': 1, ``date'': [``2024-11-21'', ``2024-11-22'', ``2024-11-23''], ``people\_number'': 6, ``local\_constraint'': {``house rule'': null, ``cuisine'': [``American'', ``Asian''], ``room type'': null, ``event'': null, ``attraction'': null, ``transportation'': null}, ``budget'': 8250.0, ``query'': null, ``level'': ``medium''\}, ``persona'': ``Traveler Type: Laidback Traveler; Purpose of Travel: Cultural Exploration; Spending Preference: Economical Traveler; Location Preference: Cities'',\textcolor{orange!80!black}{``disruption tolerance'': ``Planbound''}, ``plan'': [\{``days'': 1, ``current\_city'': ``from Miami to Louisville'', ``transportation'': ``Flight Number: F1028442, from Miami to Louisville, Departure Time: 15:29, Arrival Time: 18:13'', ``breakfast'': ``-'', ``attraction'': ``-'', ``lunch'': ``-'', ``dinner'': ``The Joy Luck, Louisville'', ``accommodation'': ``HIghlands Modern Get Away, Louisville'', ``event'': ``-'', ``point\_of\_interest\_list'': ``HIghlands Modern Get Away, stay from 19:00 to 21:00, nearest transit: Taylorsville @ Bardstown, 142.01m away;\textcolor{orange!80!black}{The Joy Luck, visit from 21:30 to 22:30, nearest transit: Bardstown @ Longest, 70.75m away};HIghlands Modern Get Away, stay from 23:00 to 08:00, nearest transit: Taylorsville @ Bardstown, 142.01m away.''\}, \{``days'': 2, ``current\_city'': ``Louisville'', ``transportation'': ``-'', ``breakfast'': ``Jack Fry's, Louisville'',\textcolor{orange!80!black}{``attraction'': ``Muhammad Ali Center,Louisville}; Frazier History Museum, Louisville; Waverly Hills Sanatorium, Louisville'', ``lunch'': ``English Grill, Louisville'', ``dinner'': ``Guy Fieri's Smokehouse, Louisville'', ``accommodation'': ``HIghlands Modern Get Away, Louisville'', ``event'': ``-'', ``point\_of\_interest\_list'': ``HIghlands Modern Get Away, stay from 08:00 to 09:00, nearest transit: Taylorsville @ Bardstown, 142.01m away;Jack Fry's, visit from 09:30 to 10:00, nearest transit: Baxter @ Highland, 14.49m away;Muhammad Ali Center, visit from 11:00 to 13:00, nearest transit: Main @ 7th, 185.29m away;English Grill, visit from 14:00 to 14:30, nearest transit: Broadway @ 4th, 59.77m away;Frazier History Museum, visit from 15:00 to 17:00, nearest transit: Market @ S, 183.51m away;Waverly Hills Sanatorium, visit from 18:00 to 20:00, nearest transit: Dixie @ Maryman, 499.50m away;Guy Fieri's Smokehouse, visit from 21:00 to 22:00, nearest transit: Muhammad Ali @ 4th, 95.01m away;HIghlands Modern Get Away, stay from 22:30 to 05:00, nearest transit: Taylorsville @ Bardstown, 142.01m away.''\}, \{``days'': 3, ``current\_city'': ``from Louisville to Miami'', ``transportation'': ``Flight Number: F1369775, from Louisville to Miami, Departure Time: 07:02, Arrival Time: 09:18'', ``breakfast'': ``Porch Kitchen \& Bar, Louisville'', ``attraction'': ``-'', ``lunch'': ``-'', ``dinner'': ``-'', ``accommodation'': ``-'', ``event'': ``-'', ``point\_of\_interest\_list'': ``HIghlands Modern Get Away, stay from 05:00 to 05:30, nearest transit: Taylorsville @ Bardstown, 142.01m away;Porch Kitchen \& Bar, visit from 06:00 to 06:30, nearest transit: W. Jefferson @ 2nd, 40.98m away.''\}]\}

\textbf{Disruption Info}\\
\textcolor{orange!80!black}{Day 2  Attraction: Muhammad Ali Center, Louisville}\\
Disruption Category: Attractions, Miscellaneous  \\
Reason: The Muhammad Ali Center is closed until further notice due to fire damage to its HVAC system, making it inaccessible to visitors. Unexpected closures can also disrupt plans for cultural exploration and cause logistical stress \\
 Disruption Detection Timestamp: 1 day before or upon arrival (via center website or local news)  \\
\textcolor{orange!80!black}{Disruption Severity: Step-level}  \\
 Confidence Level: High 

 \textbf{Revised Plan}\\
 \{``acknowledgement'': ``Attractions Disruption identified and understood. Proceeding with appropriate revisions based on disruption severity and traveler tolerance.'', ``idx'': 37, ``row\_number'': 401, ``JSON'': \{``org'': ``Miami'', ``dest'': ``Louisville'', ``days'': 3, ``visiting\_city\_number'': 1, ``date'': [``2024-11-21'', ``2024-11-22'', ``2024-11-23''], ``people\_number'': 6, ``local\_constraint'': \{``house rule'': null, ``cuisine'': [``American'', ``Asian''], ``room type'': null, ``event'': null, ``attraction'': null, ``transportation'': null     \}, ``budget'': 8250.0, ``query'': null, ``level'': ``medium''\}, ``persona'': ``Traveler Type: Laidback Traveler; Purpose of Travel: Cultural Exploration; Spending Preference: Economical Traveler; Location Preference: Cities'', ``disruption\_tolerance'': ``Planbound'', ``plan'': [    \{``days'': 1, ``current\_city'': ``from Miami to Louisville'', ``transportation'': ``Flight Number: F1028442, from Miami to Louisville, Departure Time: 15:29, Arrival Time: 18:13'', ``breakfast'': ``-'', ``attraction'': ``-'', ``lunch'': ``-'', ``dinner'': ``Eiffel Tower Restaurant, Louisville'', ``accommodation'': ``HIghlands Modern Get Away, Louisville'', ``event'': ``-'', ``point\_of\_interest\_list'': ``HIghlands Modern Get Away, stay from 19:00 to 21:00, nearest transit: Taylorsville @ Bardstown, 142.01m away;\textcolor{orange!80!black}{Eiffel Tower Restaurant, visit from 21:30 to 22:30, nearest transit: The Strat, 4118.11m away;}HIghlands Modern Get Away, stay from 23:00 to 08:00, nearest transit: Taylorsville @ Bardstown, 142.01m away.''\},     \{``days'': 2, ``current\_city'': ``Louisville'', ``transportation'': ``-'', ``breakfast'': ``Jack Fry's, Louisville'', \textcolor{orange!80!black}{``attraction'': ``Frazier History Museum, Louisville; Waverly Hills Sanatorium, Louisville; Speed Art Museum, Louisville''}, ``lunch'': ``English Grill, Louisville'', ``dinner'': ``Guy Fieri's Smokehouse, Louisville'', ``accommodation'': ``HIghlands Modern Get Away, Louisville'', ``event'': ``-'', ``point\_of\_interest\_list'': ``HIghlands Modern Get Away, stay from 08:00 to 09:00, nearest transit: Taylorsville @ Bardstown, 142.01m away;Jack Fry's, visit from 09:30 to 10:00, nearest transit: Baxter @ Highland, 14.49m away;Frazier History Museum, visit from 11:00 to 13:00, nearest transit: Market @ S, 183.51m away;English Grill, visit from 14:00 to 14:30, nearest transit: Broadway @ 4th, 59.77m away;Speed Art Museum, visit from 15:00 to 17:00, nearest transit: S. 3rd @ Cardinal, 120.45m away;Waverly Hills Sanatorium, visit from 18:00 to 20:00, nearest transit: Dixie @ Maryman, 499.50m away;Guy Fieri's Smokehouse, visit from 21:00 to 22:00, nearest transit: Muhammad Ali @ 4th, 95.01m away;HIghlands Modern Get Away, stay from 22:30 to 05:00, nearest transit: Taylorsville @ Bardstown, 142.01m away.''\},     \{``days'': 3, ``current\_city'': ``from Louisville to Miami'', ``transportation'': ``Flight Number: F1369775, from Louisville to Miami, Departure Time: 07:02, Arrival Time: 09:18'', ``breakfast'': ``Porch Kitchen \& Bar, Louisville'', ``attraction'': ``-'', ``lunch'': ``-'', ``dinner'': ``-'', ``accommodation'': ``-'', ``event'': ``-'', ``point\_of\_interest\_list'': ``HIghlands Modern Get Away, stay from 05:00 to 05:30, nearest transit: Taylorsville @ Bardstown, 142.01m away;Porch Kitchen \& Bar, visit from 06:00 to 06:30, nearest transit: W. Jefferson @ 2nd, 40.98m away.''\}   ] \} \\

\end{tcolorbox}
\end{center}

\textbf{Analysis:} We can observe the traveler has tolerance level of ``Plan-Bound"with disruption severity of Step-level, but the LLM hallucinates and struggles to identify the relation and generates a revised plan which updates multiple days POI list.

\end{document}